\documentclass{article}

    \usepackage[final]{neurips_2023}

\usepackage[utf8]{inputenc} % allow utf-8 input
\usepackage[T1]{fontenc}    % use 8-bit T1 fonts
\usepackage{hyperref}       % hyperlinks
\usepackage{url}            % simple URL typesetting
\usepackage{booktabs}       % professional-quality tables
\usepackage{amsfonts}       % blackboard math symbols
\usepackage{nicefrac}       % compact symbols for 1/2, etc.
\usepackage{microtype}      % microtypography
\usepackage{xcolor}         % colors

\usepackage{bm}
\usepackage{multirow}
\usepackage{algorithm, algpseudocode}

\usepackage{microtype}
\usepackage{graphicx}
\usepackage{subfigure}
\usepackage{booktabs} % for professional tables
\usepackage{hyperref}
\usepackage{amsmath}
\usepackage{amssymb}
\usepackage{mathtools}
\usepackage{amsthm}
\usepackage{amsfonts,bm}
\usepackage[capitalize,noabbrev]{cleveref}
\theoremstyle{plain}

\theoremstyle{definition}

\theoremstyle{remark}

\usepackage[textsize=tiny]{todonotes}

\title{Improving Robustness with Adaptive Weight Decay}

\author{%
  Amin Ghiasi, Ali Shafahi, Reza Ardekani\\
  Apple\\
  Cupertino, CA, 95014 \\
  \texttt{ \{mghiasi2, ashafahi, rardekani\} @apple.com} \\
}

\begin{document}

\newcommand{\reza}[1]{{  \textcolor{red}{\bf \em [Reza: #1]}}}
\newcommand{\rezarm}[1]{\sout {{  \textcolor{red}{\bf \em [#1]}}}}
\newcommand{\ali}[1]{}
\newcommand{\amin}[1]{{  \textcolor{green}{\bf \em [Amin: #1]}}}

\newcommand{\no}{-}
\newcommand{\yes}{\checkmark}

\maketitle

\begin{abstract}
% \amin{TODO: We should also update everything we promised we'd do during the rebuttal.}
  We propose adaptive weight decay, which automatically tunes the hyper-parameter for weight decay during each training iteration. For classification problems, we propose changing the value of the weight decay hyper-parameter on the fly based on the strength of updates from the classification loss (i.e., gradient of cross-entropy), and the regularization loss (i.e., $\ell_2$-norm of the weights). We show that this simple modification can result in large improvements in adversarial robustness — an area which suffers from robust overfitting — without requiring extra data across various datasets and architecture choices. For example, our reformulation results in 20\% relative robustness improvement for CIFAR-100, and 10\% relative robustness improvement on CIFAR-10 comparing to the best tuned hyper-parameters of traditional weight decay resulting in models that have comparable performance to SOTA robustness methods. In addition, this method has other desirable properties, such as less sensitivity to learning rate, and smaller weight norms, which the latter contributes to robustness to overfitting to label noise, and pruning.

\end{abstract}

\section{Introduction}
Deep Neural Networks (DNNs) have exceeded human capability on many computer vision tasks.
Due to their high capacity for memorizing training examples \citep{zhang2021understanding}, DNN generalization heavily relies on the training algorithm.
To reduce memorization and improve generaliazation, several approaches have been taken including regularization and augmentation. 
Some of these augmentation techniques alter the network input \citep{devries2017improved, chen2020gridmask, cubuk2019autoaugment, cubuk2020randaugment, muller2021trivialaugment}, some alter hidden states of the network \citep{srivastava2014dropout, ioffe2015batch, gastaldi2017shake, yamada2019shakedrop}, some alter the expected output \citep{warde201611, kannan2018adversarial}, and some affect multiple levels \citep{zhang2017mixup, yun2019cutmix, hendrycks2019augmix}. 
Typically, augmentation methods aim to enhance generalization by increasing the diversity of the dataset. 
The utilization of regularizers, such as weight decay \citep{plaut1986experiments, krogh1991simple}, serves to prevent overfitting by eliminating solutions that solely memorize training examples and by constraining the complexity of the DNN.
% Typically, augmentation methods attempt to improve generalization by improving the diversity of the dataset. 
% Use of regularizers, such as weight decay \citep{plaut1986experiments, krogh1991simple} prevent overfitting by eliminating solutions that memorize training examples and by limiting the complexity of the DNN. 
% Another popular approach to prevent overfitting is the use of regularizers, such as weight decay \citep{plaut1986experiments, krogh1991simple}.  
% Such methods prevent overfitting by eliminating solutions that memorize training examples and by limiting the complexity of the DNN. 
Regularization methods are most beneficial in areas such as adversarial robustness, and noisy-data settings -- settings which suffer from catastrophic overfitting. 
In this paper, we revisit weight decay; a regularizer mainly used to avoid overfitting.

The rest of the paper is organized as follows: 
In Section \ref{subsec:adv}, we revisit tuning the weight decay hyper-parameter to improve adversarial robustness and introduce Adaptive Weight Decay. Also in Section~\ref{subsec:adv}, through extensive experiments on various image classification datasets, we show that adversarial training with Adaptive Weight Decay improves both robustness and natural generalization compared to traditional non-adaptive weight decay. 
Next, in Section \ref{sec:properties}, we briefly mention other potential applications of Adaptive Weight Decay to network pruning, robustness to sub-optimal learning-rates, and training on noisy labels. 

\section{Adversarial Robustness} \label{subsec:adv}
DNNs are susceptible to adversarial perturbations \citep{szegedy2013intriguing, biggio2013evasion}. 
% It is known that Deep neural networks are susceptible to adversarial perturbations \citep{szegedy2013intriguing, biggio2013evasion}. 
In the adversarial setting, the adversary adds a small imperceptible noise to the image, which fools the network into making an incorrect prediction.
To ensure that the adversarial noise is imperceptible to the human eye, usually noise with bounded $\ell_p$-norms have been studied \citep{sharif2018suitability}. 
In such settings, the objective for the adversary is to maximize the following loss:
\begin{equation}
 \max_{|\delta|_p \leq \epsilon} Xent(f(x + \delta, w), y), \label{eq:adv_example}
\end{equation}
where $Xent$ is the Cross-entropy loss, $\delta$ is the adversarial perturbation, $x$ is the clean example, $y$ is the ground truth label, $\epsilon$ is the adversarial perturbation budget, and $w$ is the DNN paramater. 
% where $Xent$ is the Cross-entropy loss, $\delta$ is the adversarial perturbation, $x + \delta$ is the adversarial example, $y$ is the ground truth label, $\epsilon$ is the adversarial perturbation budget, and $w$ is the DNN paramater. 

A multitude of papers concentrate on the adversarial task and propose methods to generate robust adversarial examples through various approaches, including the modification of the loss function and the provision of optimization techniques to effectively optimize the adversarial generation loss functions \citep{goodfellow2014explaining, madry2017towards, carlini2017towards, izmailov2018averaging, croce2020minimally, andriushchenko2020square}.
% An array of papers focus on the adversarial task and design methods for generating strong adversarial examples by either modifying the loss function in eq~.\ref{eq:adv_example} or providing optimization tricks for optimization of the adversarial generation loss functions \citep{goodfellow2014explaining, madry2017towards, carlini2017towards, izmailov2018averaging, croce2020minimally, andriushchenko2020square}. 
An additional area of research centers on mitigating the impact of potent adversarial examples. While certain studies on adversarial defense prioritize approaches with theoretical guarantees \citep{wong2018provable, cohen2019certified}, in practical applications, variations of adversarial training have emerged as the prevailing defense strategy against adversarial attacks \citep{madry2017towards, shafahi2019adversarial, wong2020fast, rebuffi2021fixing, gowal2020uncovering}.
% Another line of work focuses on defense against strong adversarial examples. 
% While some adversarial defense studies focus on defenses with theoretical guarantees \citep{wong2018provable, cohen2019certified}, in practice, variations of adversarial training have been the de-facto defense against adversarial attacks \citep{madry2017towards, shafahi2019adversarial, wong2020fast, rebuffi2021fixing, gowal2020uncovering}. 
Adversarial training involves on the fly generation of adversarial examples during the training process and subsequently training the model using these examples.
% Adversarial training consists of on the fly generation of adversarial examples for training data and training on them. 
The adversarial training loss can be formulated as a min-max optimization problem:
% \vspace{-4em}
\begin{equation}
  \min_{w} \max_{|\delta|_p \leq \epsilon}  Xent(f(x + \delta, w), y)   , \label{eq:adv_training_only}
\end{equation}
% \vspace{-6em}

\subsection{Robust overfitting and relationship to weight decay}\label{subsec:robust_overfit_weightdecay}
Adversarial training is a strong baseline for defending against adversarial attacks; however, it often suffers from a phenomenon referred to as \emph{Robust Overfitting} \citep{rice2020overfitting}.
Weight decay regularization, as discussed in \ref{subsec:wd}, is a common technique used for preventing overfitting.

\subsubsection{Weight Decay} \label{subsec:wd} % [ ]
Weight decay encourages weights of networks to have smaller magnitudes \citep{zhang2018three} and is widely used to improve generalization. 
Weight decay regularization can have many forms \citep{loshchilov2017decoupled}, and we focus on the popular $\ell_2$-norm variant.
More precisely, we focus on classification problems with cross-entropy as the main loss -- such as adversarial training -- and weight decay as the regularizer, which was popularized by \cite{krizhevsky2017imagenet}: 

\begin{equation}
% \vspace{-4em}
 Loss_w(x, y) = Xent(f(x, w),y) +\frac{ \lambda_{wd} }{2} \|w\|_2^2, \label{eq:reg_wd}
 % \vspace{-4em}
\end{equation}

% where $w$ is the network parameters, ($x$, $y$) is the training data, and $\lambda_{wd}$ is the hyper-parameter controlling how much weight decay penalizes the norm of weights compared to the main loss (i.e., cross-entropy loss).
where $w$ is the network parameters, ($x$, $y$) is the training data, and $\lambda_{wd}$ is the weight-decay hyper-parameter. 
% If $\lambda_{wd}$ is too small, the optimization may overfit the data, while a value that is too large may result in a solution with a low weight-norm that does not accurately fit the training data. 
$\lambda_{wd}$ is a crucial hyper-parameter in weight decay, determining the weight penalty compared to the main loss (e.g., cross-entropy). A small $\lambda_{wd}$ may cause overfitting, while a large value can yield a low weight-norm solution that poorly fits the training data. 
Thus, selecting an appropriate $\lambda_{wd}$ value is essential for achieving an optimal balance.

% \vspace{2mm}
\subsubsection{Robust overfitting phenomenon revisited}\label{subsubsec:robust_overfit} % [ ]
% \vspace{2mm}

To study robust overfitting, we focus on evaluating the $\ell_{\infty}$ adversarial robustness on the CIFAR-10 dataset while limiting the adversarial budget of the attacker to $\epsilon=8$ -- a common setting for evaluating robustness. 
For these experiments, we use a WideResNet 28-10 architecture \citep{zagoruyko2016wide} and widely adopted PGD adversarial training \citep{madry2017towards} to solve the adversarial training loss with weight decay regularization: 

\begin{equation}
  \min_{w} \big(\max_{|\delta|_{\infty} \leq 8}  Xent(f(x + \delta, w), y) + \frac{\lambda_{wd}}{2} \|w\|^2_2  \big), \label{eq:adv_training}
\end{equation}

We reserve 10\% of the training examples as a held-out validation set for early stopping and checkpoint selection. 
In practice, to solve eq.~\ref{eq:adv_training}, the network parameters $w$ are updated after generating adversarial examples in real-time using a 7-step PGD adversarial attack. 
We train for 200 epochs, using an initial learning-rate of 0.1 combined with a cosine learning-rate schedule. 
Throughout training, at the end of each epoch, the robust accuracy and robustness loss (i.e., cross-entropy loss of adversarial examples) are evaluated on the validation set by subjecting the held-out validation examples to a 3-step PGD attack.
% During training, after each epoch, we evaluate the robust accuracy and the robustness loss (i.e. cross-entropy loss of adversarial examples) on the validation set by conducting a 3-step PGD attack. % on the validation examples.
For further details, please refer to \ref{app:setup_adv}.

To further understand the robust overfitting phenomenon in the presence of weight decay, we train different models by varying the weight-norm hyperparameter $\lambda_{wd}$ in eq.~\ref{eq:adv_training}. 

\begin{figure}[pt]
    % \vspace{-3em}
    \centering
    \subfigure[]{\includegraphics[width=0.31\columnwidth]{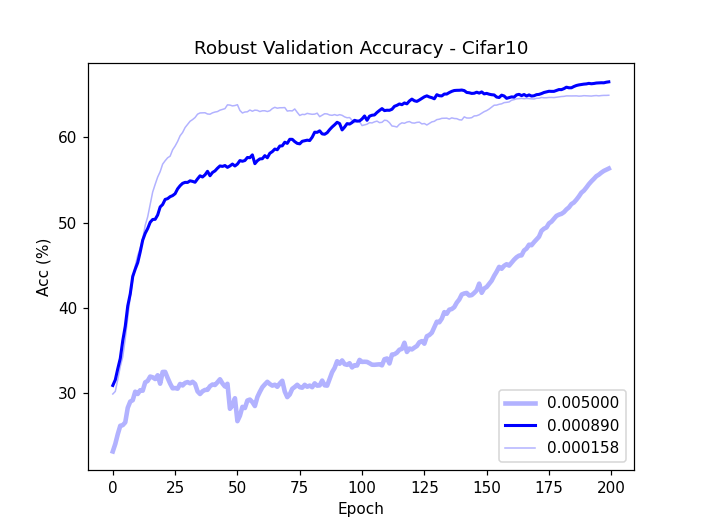}\label{fig:adv_val_acc_c10}}
    \subfigure[]{\includegraphics[width=0.31\columnwidth]{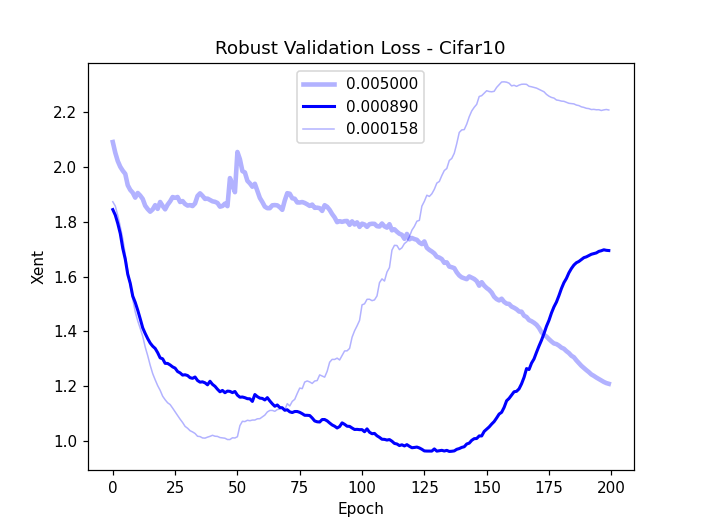}\label{fig:adv_val_loss_c10}}
    \subfigure[]{\includegraphics[width=0.31\columnwidth]{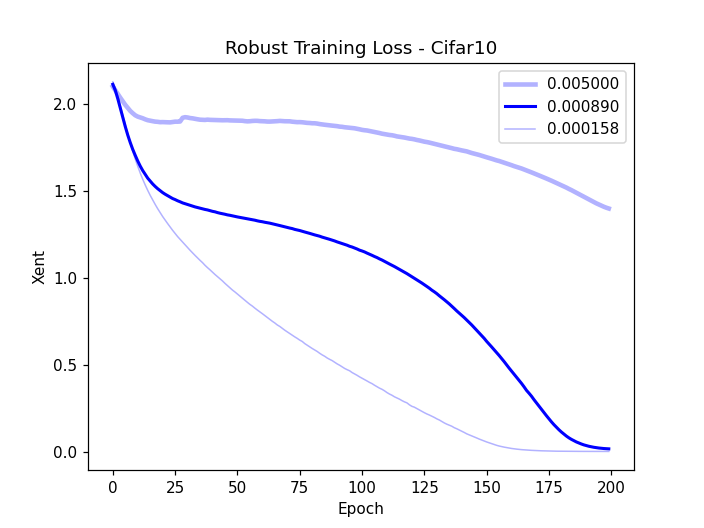}\label{fig:adv_train_loss_c10}}
    \caption{ Robust validation accuracy (a) and validation loss (b) and training loss (c) on CIFAR-10 subsets. $\lambda_{wd} = 0.00089$ is the best performing hyper-parameter we found by doing a grid-search. The other two hyper-parameters are two points from our grid-search, one with larger and the other with smaller hyper-parameter for weight decay. The thickness of the plot-lines correspond to the magnitude of the weight-norm penalties. As it can be seen by (a) and (b), networks trained by small values of $\lambda_{wd}$ suffer from robust-overfitting, while networks trained with larger values of $\lambda_{wd}$ do not suffer from robust overfitting but the larger $\lambda_{wd}$ further prevents the network from fitting the data (c) resulting in reduced overall robustness. }
    \label{fig:adv_over_fitting}
    \vspace{-1em}
\end{figure}

Figure~\ref{fig:adv_over_fitting} illustrates the accuracy and cross-entropy loss on the adversarial examples built for the held-out validation set for three choices\footnote{Figure~\ref{fig:adv_over_fitting_awd} captures the complete set of $\lambda_{wd}$ values we tested.} of $\lambda_{wd}$ throughout training.
As seen in Figure~\ref{fig:adv_val_acc_c10}, for small $\lambda_{wd}$ choices, the robust validation accuracy does not monotonically increase towards the end of training. 
The Non-monotonicity behavior, which is related to robust overfitting, is even more pronounced if we look at the robustness loss computed on the held-out validation (Figure~\ref{fig:adv_val_loss_c10}). 
Note that this behavior is still evident even if we look at the best hyper-parameter value according to the validation set ($\lambda_{wd} ^ * = 0.00089$). 

Various methods have been proposed to rectify robust overfitting, including early stopping \citep{rice2020overfitting}, use of extra unlabeled data \citep{carmon2019unlabeled},  synthesized images \citep{gowal2020uncovering}, pre-training \citep{hendrycks2019using}, use of data augmentations \citep{rebuffi2021fixing}, and stochastic weight averaging \citep{izmailov2018averaging}. 

% Interestingly, we observe that having smaller weight-norms (by increasing $\lambda_{wd}$) could reduce this non-monotonic behavior on the validation set adversarial examples. 
In Fig.~\ref{fig:adv_over_fitting}, we observe that simply having smaller weight-norms (by increasing $\lambda_{wd}$) could reduce this non-monotonic behavior on the validation set adversarial examples. 
Although, this comes at the cost of larger cross-entropy loss on the training set adversarial examples, as shown in Figure~\ref{fig:adv_train_loss_c10}. 
Even though the overall loss function from eq.~\ref{eq:adv_training} is a minimization problem, the terms in the loss function implicitly have conflicting objectives: 
During the training process, when the cross-entropy term holds dominance, effectively reducing the weight norm becomes challenging, resulting in non-monotonic behavior of robust validation metrics towards the later stages of training. Conversely, when the weight-norm term takes precedence, the cross-entropy objective encounters difficulties in achieving significant reductions.
% During training when the cross-entropy term dominates, it is hard to reduce the norm of the weights, and the robust validation metrics have non-monotonic behavior towards the end of training. On the other end of the spectrum, when the weight-norm term dominates, the cross-entropy objective struggles to reduce. 
In the next section, we introduce \emph{Adaptive Weight Decay}, which explicitly strikes a balance between these two terms during training. 

\subsection{Adaptive Weight Decay} 
Inspired by the findings in \ref{subsubsec:robust_overfit}, we propose \textbf{A}daptive \textbf{W}eight \textbf{D}ecay (AWD). 
The goal of AWD is to maintain a balance between weight decay and cross-entropy updates during training in order to guide the optimization to a solution which satisfies both objectives more effectively. 
To derive AWD, we study one gradient descent step for updating the parameter $w$ at step $t+1$ from its value at step $t$:
\begin{equation}
 w_{t+1} = w_{t} - \nabla w_{t} \cdot lr - w_{t} \cdot \lambda_{wd} \cdot lr, \label{eq:gd_update}
\end{equation}
where $\nabla w_{t}$ is the gradient computed from the cross-entropy objective, and $w_t\cdot\lambda_{wd}$ is the gradient computed from the weight decay term from eq.~\ref{eq:reg_wd}. 
% The first term ($lr \cdot \nabla w_{t}$) comes from the contribution of the cross-entropy loss, while the second term ($lr \cdot \lambda_{wd} \cdot \|w\|_2$) is the contribution from weight decay. 
% We define \textbf{D}ecay \textbf{o}ver \textbf{G}radient of cross-entropy (DoG for short) as a metric that keeps track of the ratio of the magnitudes coming from each objective: 
We define \textbf{$\lambda_{awd}$} as a metric that keeps track of the ratio of the magnituedes coming from each objective:
\begin{equation}
 % DoG_t=\frac{\| \lambda_{wd} w_t\|}{\|\nabla w_t\|}, \label{eq:dog}
 {\lambda_{awd}}_{(t)}=\frac{\| \lambda_{wd} w_t\|}{\|\nabla w_t\|}, \label{eq:dog}
\end{equation}

To keep a balance between the two objectives, we aim to keep this ratio constant during training. 
AWD is a simple yet effective way of maintaining this balance. % during training. 
Adaptive weight decay shares similarities with non-adaptive (traditional) weight decay, with the only distinction being that the hyper-parameter $\lambda_{wd}$ is not fixed throughout training. Instead, $\lambda_{wd}$ dynamically changes in each iteration to ensure ${\lambda_{awd}}_{(t)} \approx {\lambda_{awd}}_{(t-1)} \approx {\lambda_{awd}}$. %$DoG_{t} \approx DoG_{t-1} \approx DoG$. 
% Adaptive weight decay is similar to non-adaptive (i.e., traditional) weight decay.
% The only difference is that the hyper-parameter $\lambda_{wd}$ is not constant throughout training and changes in every iteration to ensure that $DoG_{t} \approx DoG_{t-1} \approx DoG$. 
% To keep this ratio constant at every step $t$, we can rewrite the $DoG$ equation (eq.~\ref{eq:dog}) as:
To keep this ratio constant at every step $t$, we can rewrite the ${\lambda_{awd}}$ equation (eq.~\ref{eq:dog}) as:

\begin{equation}
 \lambda_{wd(t)} = \frac{{\lambda_{awd}} \cdot \| \nabla w_t \| }{\| w_t\|}, \label{eq:awd}
\end{equation}

Eq.~\ref{eq:awd} allows us to have a different weight decay hyperparameter value ($\lambda_{wd}$) for every optimization iteration $t$, which keeps the gradients received from the cross entropy and weight decay balanced throughout the optimization. 
Note that weight decay penalty $\lambda_t$ can be computed on the fly with almost no computational overhead during the training. 
% Using our approach, the value of $\lambda_{t}$ varies in each iteration.
% However, using the exponential weighted average $\bar{\lambda_{t}} = 0.1 \times \bar{\lambda_{t-1}} + 0.9 \times \lambda_t$  could make $\lambda_{t}$ more stable. 
Using the exponential weighted average $\bar{\lambda_{t}} = 0.1 \times \bar{\lambda_{t-1}} + 0.9 \times \lambda_t$, we could make $\lambda_{t}$ more stable (Algorithm \ref{app:alg}). 
% Algorithm \ref{app:alg} provides a pseudo-code of our implementation. 

\begin{algorithm}
\caption{Adaptive Weight Decay}\label{app:alg}
\begin{algorithmic}[1]
\State {\bfseries Input:} ${\lambda_{awd}} > 0$
% \Require ${\lambda_{awd}} > 0$
\State $\bar{\lambda} \gets 0$ 
\For{$(x,y) \in loader$}
\State $p \gets model(x)$ \hfill\Comment{Get models prediction.}
\State $main \gets CrossEntropy(p, y)$ \hfill\Comment{Compute CrossEntropy.}
\State $\nabla w \gets backward(main)$ \hfill\Comment{Compute the gradients of main loss w.r.t weights.}
\State $\lambda \gets \frac{\| \nabla w \| {\lambda_{awd}}}{\| w \|}$ \hfill\Comment{Compute iteration's weight decay hyperparameter.}
\State $\bar{\lambda} \gets 0.1 \times \bar{\lambda} + 0.9 \times stop\_gradient(\lambda) $ \hfill\Comment{Compute the weighted average as a scalar.}
\State $w \gets w - lr( \nabla w + \bar{\lambda} \times w ) $ \hfill\Comment{Update Network's parameters.}
\EndFor
\end{algorithmic}
\end{algorithm}

\subsubsection{Differences between Adaptive and Non-Adaptive Weight Decay} \label{sec:awd_diff_trad}
To study the differences between adaptive and non-adaptive weight decay and to build intuition, we can plug in  $\lambda_t$ of the adaptive method (eq.~\ref{eq:awd}) directly into the equation for traditional weight decay (eq.~\ref{eq:reg_wd}) and derive the total loss based on Adaptive Weight Decay: 
\begin{equation}
 Loss_{w_t}(x, y) = Xent(f(x, w_t),y) + \frac{{\lambda_{awd}} \cdot \| \nabla w_t \| \|w_t\|}{2}, \label{eq:awd_weight_decay}
\end{equation}

Please note that directly solving eq.~\ref{eq:awd_weight_decay} will invoke the computation of second-order derivatives since $\lambda_t$ is computed using the first-order derivatives. 
However, as stated in Alg.~\ref{app:alg}, we convert the $\lambda_{t}$ into a non-tensor scalar to save computation and avoid second-order derivatives. 
We treat $\|\nabla w_t\|$ in eq.~\ref{eq:awd_weight_decay} as a constant and do not allow gradients to back-propagate through it. 
As a result, adaptive weight decay has negligible computation overhead compared to traditional non-adaptive weight decay. 

By comparing the weight decay term in the adaptive weight decay loss (eq.~\ref{eq:awd_weight_decay}): $\frac{{\lambda_{awd}}}{2} \|w\| \|\nabla w\|$ with that of the traditional weight decay loss (eq.~\ref{eq:reg_wd}): $\frac{\lambda_{wd}}{2} \|w\|^2$, we can build intuition on some of the differences between the two.  
For example, the non-adaptive weight decay regularization term approaches zero only when the weight norms are close to zero, whereas, in AWD, it also happens when the  cross-entropy gradients are close to zero. 
Consequently, AWD prevents over-optimization of weight norms in flat minima, allowing for more (relative) weight to be given to the cross-entropy objective.
Additionally, AWD penalizes weight norms more when the gradient of cross-entropy is large, preventing it from falling into steep local minima and hence overfitting early in training.

We verify our intuition of AWD being capable of reducing robust overfitting in practice by replacing the non-adaptive weight decay with AWD and monitoring the same two metrics from \ref{subsubsec:robust_overfit}. The results for a good choice of the AWD hyper-parameter (${\lambda_{awd}}$) and various choices of non-adaptive weight decay ($\lambda_{wd}$) hyper-parameter are summarized in Figure~\ref{fig:adv_over_fitting_awd} \footnote{See Appendix \ref{app:c100_robust_overf} for similar analysis on other datasets.}. %CIFAR-100
% As can be seen, when we train with adaptive weight decay, we are able to reduce robust overfitting. Also, the best models trained with adaptive weight decay have considerably smaller weight-norms compared to the best models trained with non-adaptive weight decay as we will show in \ref{subsec:results_robustness}. 

% \subsubsection{Reducing Robust Overfitting using Adaptive Weight Decay} % [#]
% When we replace non-adaptive weight decay with AWD and monitor the same two metrics from \ref{subsubsec:robust_overfit}, we can see that AWD suffers less from robust overfitting. The results for a good choice of the adaptive weight decay hyper-parameter ($DoG$) and various choices of non-adaptive weight decay ($\lambda_{wd}$) are summarized in Figure~\ref{fig:adv_over_fitting_awd} \footnote{See Appendix \ref{app:c100_robust_overf} for similar analysis on other datasets.}. %CIFAR-100
% As can be seen, when we train with adaptive weight decay, we are able to reduce robust overfitting. Also, the best models trained with adaptive weight decay have considerably smaller weight-norms compared to the best models trained with non-adaptive weight decay as we will show in \ref{subsec:results_robustness}. 

% \amin{Figure -- similar to before but with all lambda choices + dog}

\begin{figure*}[t]
    \vspace{-1em}
    \centering
    \subfigure[]{\includegraphics[width=0.49\columnwidth]{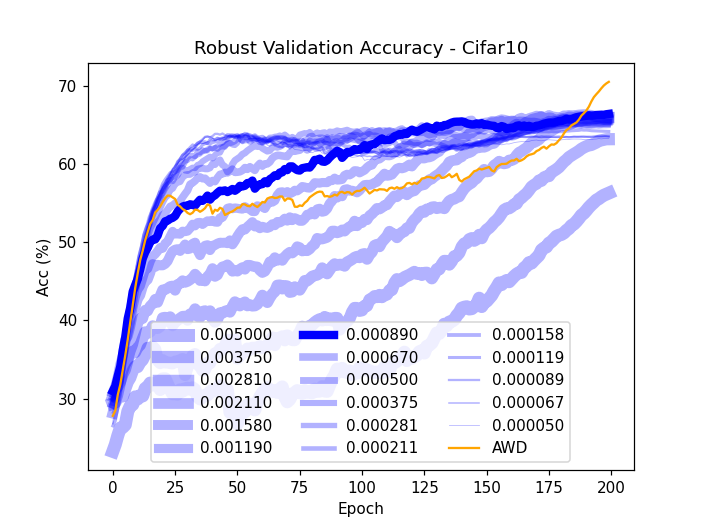}\label{fig:adv_val_acc_c10_all}}
    \subfigure[]{\includegraphics[width=0.49\columnwidth]{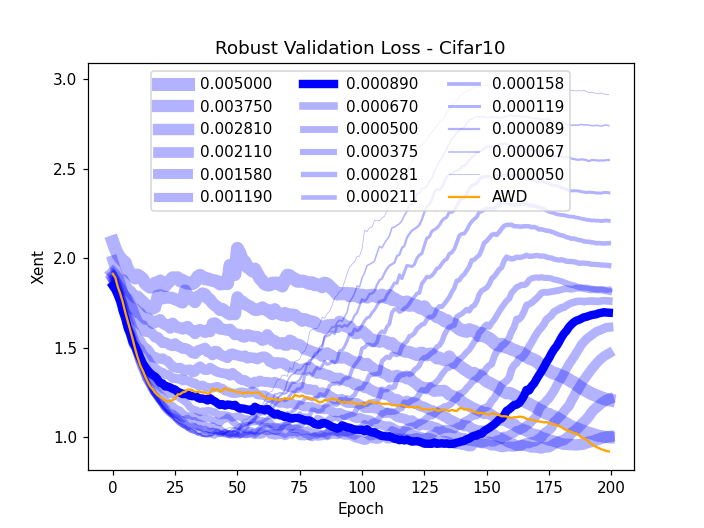}\label{fig:adv_val_loss_c10_all}}
    \caption{ Robust accuracy (a) and loss (b) on CIFAR-10 validation subset. Both figures highlight the best performing hyper-parameter for non-adaptive weight decay $\lambda_{wd} = 0.00089$ with sharp strokes. As it can be seen, lower values of $\lambda_{wd}$ cause robust overfitting, while high values of it prevent network from fitting entirely. However, training with adaptive weight decay prevents overfitting and achieves highest performance in robustness. }
    \label{fig:adv_over_fitting_awd}
    \vspace{-1em}
\end{figure*}

\subsubsection{Related works to Adaptive Weight Decay} 
% \ali{new subsection for moving the 2 very related methods to the paper}

The most related studies to AWD are \emph{AdaDecay} \citep{nakamura2019adaptive} and \emph{LARS} \citep{you2017large}.
% AdaDecay is the first method introducing the concept of adaptive weight decay. 
% AdaDecay changes the weight decay hyper-parameter adaptively for each individual parameter, as opposed to ours which we tune the hyper-parameter for the entire network. 
AdaDecay changes the weight decay hyper-parameter adaptively for each individual parameter, as opposed to ours which we tune the hyper-parameter for the entire network. 
LARS is a common optimizer when using large batch sizes which adaptively changes the learning rate for each layer. 
We evaluate these relevant methods in the context of improving adversarial robustness and experimentally compare with AWD in Table~\ref{tab:sota} and Appendix~\ref{app:related_AWD} \footnote{Due to space limitations we defer detailed discussions and comparisons to Appendix~\ref{app:related_AWD}. 
}. 
% We also discuss AWDs mathematical differences to both methods and more detailed discussion/comparison in Appendix~\ref{app:related_AWD}. Table~\ref{tab:sota} shows the performance for the best hyper-parameter found for AdaDecay. The experimental comparisons for LARS and more detailed discussion on both methods can be found in Appendix~\ref{app:related_AWD}.

\subsection{Experimental Robustness results for Adaptive Weight Decay}\label{subsec:results_robustness}
AWD can help improve the robustness on various datasets which suffer from robust overfitting.
To illustrate this, we focus on six datasets: SVHN, FashionMNIST, Flowers, CIFAR-10, CIFAR-100, and Tiny ImageNet. 
Tiny ImageNet is a subset of ImageNet, consisting of 200 classes and images of size $64\times64\times3$. 
For all experiments, we use the widely accepted 7-step PGD adversarial training to solve eq.~\ref{eq:adv_training} \citep{madry2017towards} while keeping 10\% of the examples from the training set as held-out validation set for the purpose of early stopping. 
For early stopping, we select the checkpoint with the highest $\ell_\infty = 8$ robustness accuracy measured by a 3-step PGD attack on the held-out validation set. 
For CIFAR10, CIFAR100, and Tiny ImageNet experiments, we use a WideResNet 28-10 architecture, and for SVHN, FashionMNIST, and Flowers, we use a ResNet18 architecture. 
Other details about the experimental setup can be found in Appendix~\ref{app:setup_adv}. 
For all experiments, we tune the conventional non-adaptive weight decay parameter ($\lambda_{wd}$) for improving robustness generalization and compare that to tuning the ${\lambda_{awd}}$ hyper-parameter for adaptive weight decay. 
To ensure that we search for enough values for $\lambda_{wd}$, we use up to twice as many values for $\lambda_{wd}$ compared to ${\lambda_{awd}}$. 

\begin{figure*}[t]
    % \vspace{-3em}
    \centering
    % \subfigure[]{\includegraphics[width=0.64\linewidth]{adv_wd_c100}\label{fig:adv_wd_c10}}
    % \subfigure[]{\includegraphics[width=0.32\linewidth]{adv_dog_c100}\label{fig:adv_dog_c10}}
    % \subfigure[]{\includegraphics[width=0.64\linewidth]{adv_wd_c10}\label{fig:adv_wd_c100}}
    % \subfigure[]{\includegraphics[width=0.32\linewidth]{adv_dog_c10}\label{fig:adv_dog_c100}}
    % \subfigure[]{\includegraphics[width=0.64\linewidth]{adv_wd_tiny}\label{fig:adv_wd_tiny}}
    % \subfigure[]{\includegraphics[width=0.32\linewidth]{adv_dog_tiny}\label{fig:adv_dog_tiny}}
    \subfigure[]{\includegraphics[width=0.64\linewidth]{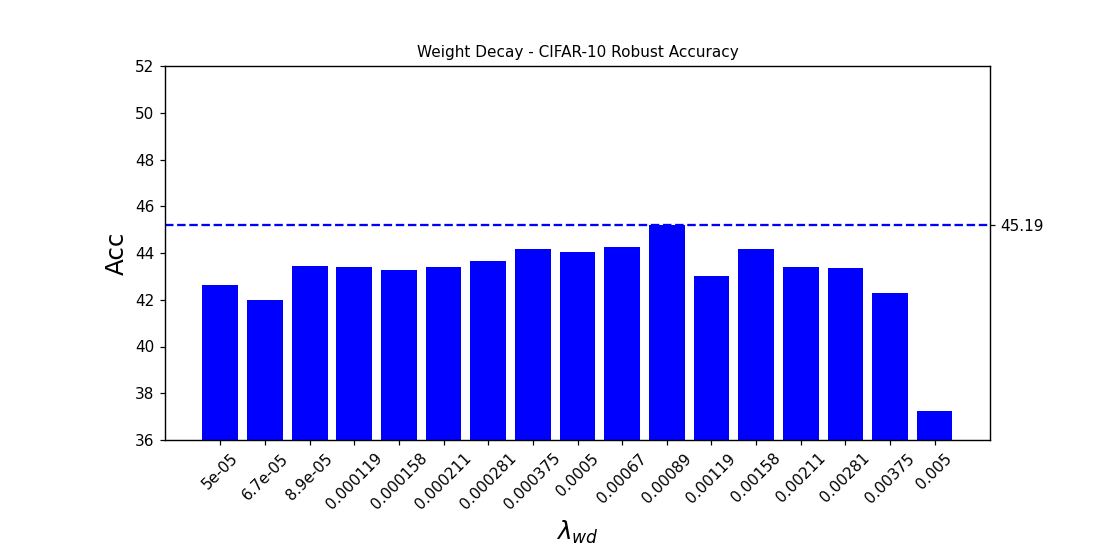}\label{fig:adv_wd_c10}}
    \subfigure[]{\includegraphics[width=0.32\linewidth]{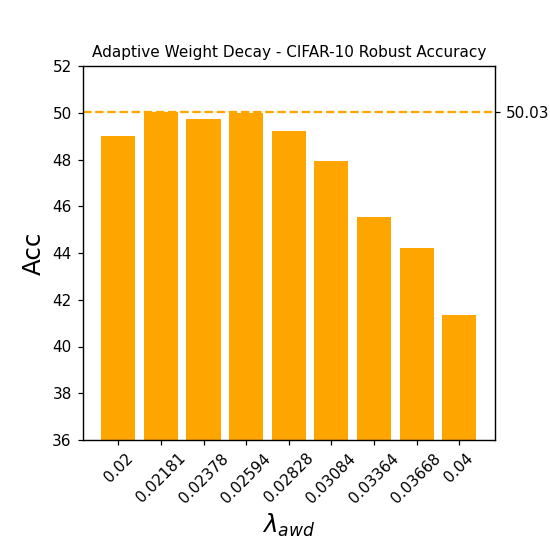}\label{fig:adv_dog_c10}}
    \subfigure[]{\includegraphics[width=0.64\linewidth]{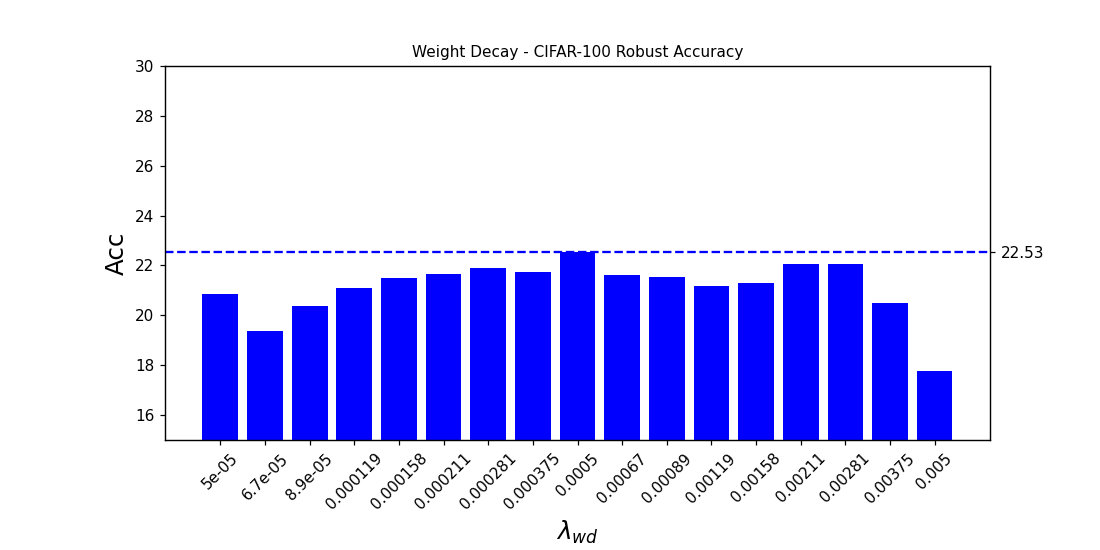}\label{fig:adv_wd_c100}}
    \subfigure[]{\includegraphics[width=0.32\linewidth]{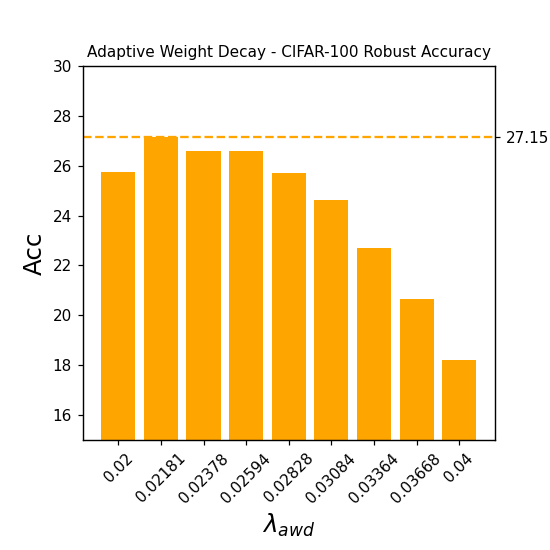}\label{fig:adv_dog_c100}}
    \subfigure[]{\includegraphics[width=0.64\linewidth]{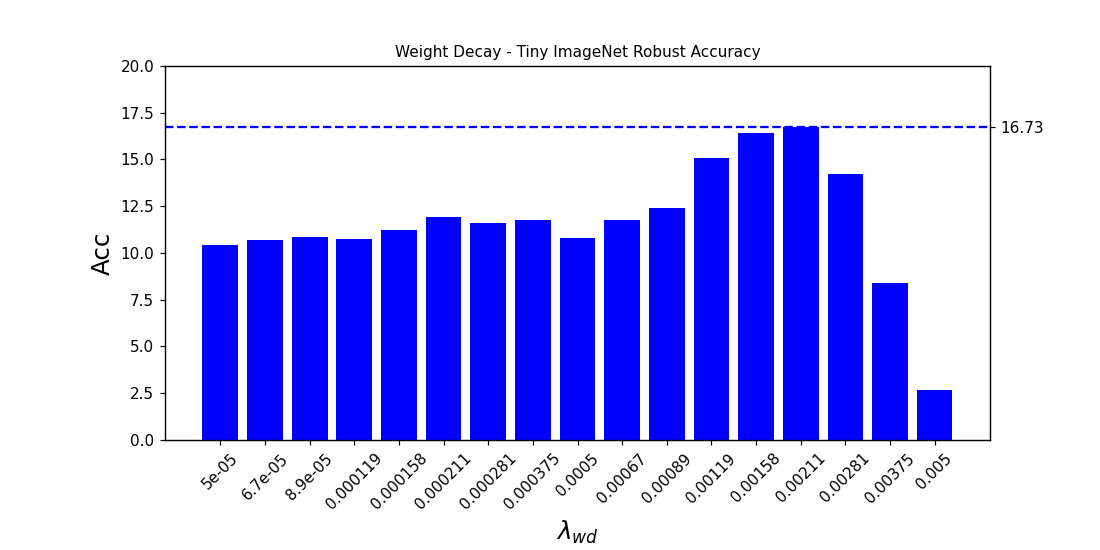}\label{fig:adv_wd_tiny}}
    \subfigure[]{\includegraphics[width=0.32\linewidth]{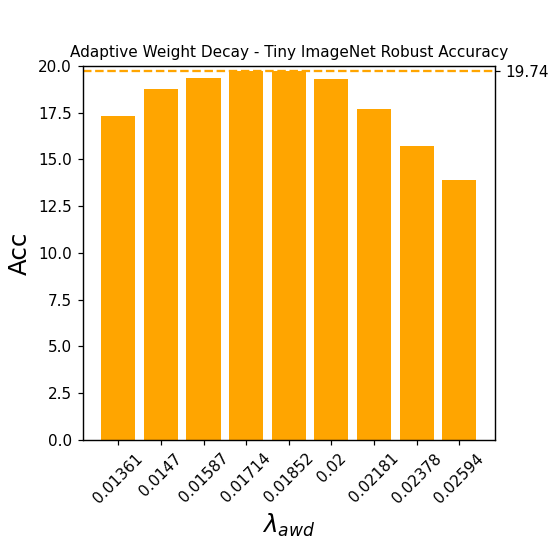}\label{fig:adv_dog_tiny}}
    \caption{ $\ell_\infty=8$ robust accuracy on the test set of adversarially trained WideResNet28-10 networks on CIFAR-10, CIFAR-100, and Tiny ImageNet (a, c, e) using different choices for the hyper-parameters of non-adaptive weight decay ($\lambda_{wd}$), and (b, d, f) different choices of the hyper-parameter for adaptive weight decay (${\lambda_{awd}}$). }
    % \ali{if we are struggling with space, we might be able to consider moving one or two rows to the appendix.}}
    \label{fig:adv}
    \vspace{-1em}
\end{figure*}

Figure~\ref{fig:adv} plots the robustness accuracy measured by applying AutoAttack \citep{croce2020reliable} on the test examples for the CIFAR-10, CIFAR-100, and Tiny ImageNet datasets, respectively. 
We observe that training with adaptive weight decay improves the robustness by a margin of 4.84\% on CIFAR-10, 5.08\% on CIFAR-100, and 3.01\% on Tiny ImageNet, %, 2.94\% on SVHN, 0.5\% on FashionMNIST, and 6.75\% on Flowers, 
compared to the non-adaptive counterpart. 
These margins translate to a relative improvement  of 10.7\%, 20.5\%, and 18.0\%, %, 6.6\%, 0.6\%, 21.2\% 
on CIFAR-10, CIFAR-100, and Tiny ImageNet, % , SVHN, FashionMNIST, and Flowers, 
respectively. 

Increasing robustness often comes at the cost of drops in clean accuracy \citep{zhang2019theoretically}. 
This observation could be attributed, at least in part, to the phenomenon that certain $\ell_p$-norm bounded adversarial examples bear a closer resemblance to the network's predicted class than their original class \citep{sharif2018suitability}.
% This could be partially due to the fact that some $\ell_p$-norm bounded adversarial examples look more like the network's prediction than their original class \citep{sharif2018suitability}.
An active area of research seeks a better trade-off between robustness and natural accuracy by finding other points on the Pareto-optimal curve of robustness and accuracy. 
For example, \citep{balaji2019instance} use instance-specific perturbation budgets during training. 
Interestingly, when comparing the most robust network trained with non-adaptive weight decay ($\lambda_{wd}$) to that trained with AWD (${\lambda_{awd}}$), we notice that those trained with the adaptive method have higher clean accuracy across various datasets (Table.~\ref{tab:nat_perf_adv_trained}). 

In addition, we observe comparatively smaller weight-norms for models trained with adaptive weight decay, which might contribute to their better generalization and lesser robust overfitting. 
For the SVHN dataset, the bes model trained with AWD has $\approx 20x$ smaller weight-norm compared to the best model trained with traditional weight-decay. 
Networks which have such small weight-norms that maintain good performance on validation data is difficult to achieve with traditional weight-decay as previously illustrated in sec.~\ref{subsubsec:robust_overfit}.
Perhaps most interestingly, when we compute the value of non-adaptive weight decay loss for AWD trained models, we observe that they are even sometimes superior in terms of that objective as it can be seen in the last column of Table~\ref{tab:nat_perf_adv_trained}. 
This behavior could imply that the AWD reformulation is better in terms of optimization and can more easily find a balance and simultaneously decrease both objective terms in eq.~\ref{eq:adv_training} and is aligned with the intuitive explanation in sec.~\ref{sec:awd_diff_trad}. 

\begin{table*}[t]
    \vspace{-1.2em}
	\centering
	\begin{tabular}{rccccccc}
    	\toprule
    	Method &						Dataset&	Opt & 	$\|W\|_2$ & 	Nat Acc &		AutoAtt & $Xent + \frac{\lambda_{wd}^* \cdot \|W\|_2^2}{2}$ \\     	
        \midrule
        
    	$\lambda_{wd} = 0.00089$ 	&	\multirow{2}{*}{CIFAR-10} &	SGD & 			35.58 & 		84.31 &  		45.19 & 0.58 \\ 
    	${\lambda_{awd}}=0.022$ &							&	SGD & 			\textbf{7.11} & 		\textbf{87.08} &			\textbf{50.03} & \textbf{0.08} \\ 
    	\midrule
     
    	$\lambda_{wd} = 0.0005$ 	&	\multirow{2}{*}{CIFAR-100} &	SGD & 			51.32 & 		60.15 &  		22.53 & \textbf{0.67} \\ 
    	${\lambda_{awd}}=0.022$ &							&	SGD & 			\textbf{13.41} & 		\textbf{61.39} &			\textbf{27.15} & 1.51 \\   
    	\midrule
     
    	$\lambda_{wd} = 0.00211$ 	&	\multirow{2}{*}{Tiny ImgNet} &	SGD & 			25.62 & 		47.87 &  		16.73 & 3.56 \\ 
    	${\lambda_{awd}}=0.01714$ &							&	SGD & 			\textbf{15.01} & 		\textbf{48.46} &			\textbf{19.74} & \textbf{2.80} \\  
        \midrule
     
        $\lambda_{wd} = 5e-7$ 	&	\multirow{2}{*}{SVHN} &	SGD & 			102.11 & 		92.04 &  		44.16 & \textbf{1.02} \\ 
    	${\lambda_{awd}}=0.02378$ &							&	SGD & 			\textbf{5.39} & 		\textbf{93.04} &			\textbf{47.10} & 1.15 \\     
        \midrule
        
    	$\lambda_{wd} = 0.00089$ 	&	\multirow{2}{*}{FashionMNIST} &	SGD & 			14.39 & 		83.96 &  		78.73 & 0.51 \\ 
    	${\lambda_{awd}}=0.01414$ &							&	SGD & 			\textbf{9.05} & 		\textbf{85.42} &			\textbf{79.24} & \textbf{0.44} \\     
    	\midrule

        $\lambda_{wd} = 0.005$ 	&	\multirow{2}{*}{Flowers} &	SGD & 			19.94 & 		\textbf{90.98} &  		32.35 & 1.72 \\ 
    	${\lambda_{awd}}=0.06727$ &							&	SGD & 			\textbf{13.87} & 		90.39 &			\textbf{39.22} & \textbf{1.42} \\     
     
		\bottomrule
    \end{tabular}
	\caption{Adversarial robustness of PGD-7 adversarially trained networks using adaptive and non-adaptive weight decay. Table summarizes the best performing hyper-parameter for each method on each dataset. Not only the adaptive method outperforms the non-adaptive method in terms of robust accuracy, it is also superior in terms of the natural accuracy. Models trained with AWD have considerably smaller weight-norms. In the last column, we report the total loss value of the non-adaptive weight decay for the best tuned $\lambda$ for that dataset, found by grid search for each dataset. Interestingly, when we measure the non-adaptive total loss (eq.~\ref{eq:adv_training}) on the training set, we observe that networks trained with the adaptive method often have smaller non-adaptive training loss even though in AWD we have not optimized that loss directly.}
	\label{tab:nat_perf_adv_trained}
 \vspace{-1.2em}
\end{table*}

The previously shown results suggest an excellent potential for adversarial training with AWD. 
To further study this potential, we only substituted the Momentum-SGD optimizer used in all previous experiments with the ASAM optimizer \citep{kwon2021asam}, and used the same hyperparameters used in previous experiments for comparison with advanced adversarial robustness algorithms. 
To the best of our knowledge, and according to the RobustBench \citep{croce2020robustbench}, the state-of-the-art $\ell_{\infty}=8.0$ defense for CIFAR-100 without extra synthesized or captured data using WRN28-10 achieves 29.80\% robust accuracy \citep{rebuffi2021fixing}. 
We achieve \emph{29.54\%} robust accuracy, which is comparable to these advanced algorithms even though our approach is a \emph{simple modification to weight decay}. % \ali{expanding table one here and referecning the architecture experiment...}
This is while our proposed method achieves 63.93\% natural accuracy which is $\approx 1\%$ higher.  
See Table~\ref{tab:sota} for more details. 
In addition to comparing with the SOTA method on WRN28-10, we compare with a large suite of strong robustness methods which report their performances on WRN32-10 in Table~\ref{tab:sota}. 
In addition to these two architectures, in Appendix~\ref{app:arch}, we test other architectures to demonstrate the robustness of the proposed method to various architectural choices. 
For ablations demonstrating the robustness of AWD to various other parameter choices for adversarial training such as number of epochs, adversarial budget ($\epsilon$), please refer to Appendix~\ref{app:epoch_vary} and Appendix~\ref{app:various_eps}, respectively. 

\begin{table*}[h]
    % \vspace{-1em}
    \hspace*{-2em}
	\centering
	\begin{tabular}{rccccccccc}
    	\toprule
    	Method &                    WRN &   Aug &   Epo &    ASAM &  TR & SWA  &  Nat &   AA \\ 
    	\midrule 
    	$\lambda_{AdaDecay} = 0.002$* & 28-10 & P\&C &  200 &   \no &  \no &   \no &      57.17 & 24.18 \\ 
    	\citep{rebuffi2021fixing} & 28-10 & CutMix & 400 &  \no &   \yes &  \yes &     62.97 & 	\textbf{29.80} \\ 
    	\citep{rebuffi2021fixing} & 28-10 & P\&C &  400 &   \no &   \yes &  \yes &      59.06 & 	28.75 \\ 
         $\lambda_{wd} = 0.0005$ &   28-10 & P\&C &  200 &   \no &   \no &   \no &      60.15 &  	22.53 \\ 
    	$\lambda_{wd} = 0.0005$ + ASAM & 	28-10 & P\&C &  100 &   \yes &  \no &   \no &     58.09 &  	22.55 \\ 
    	$\lambda_{wd} = 0.00281$* + ASAM & 28-10 & P\&C &  100 &   \yes &  \no &   \no &      62.24 &  	26.38 \\
        $\bm{\lambda_{awd}} = 0.022$  &	            28-10 & P\&C &  200 &   \no &   \no &   \no &      61.39 &		27.15 \\ 
    	$\bm{\lambda_{awd}} = 0.022$ + ASAM &				28-10 & P\&C &  100 &   \yes &  \no &   \no &      \textbf{63.93} &	    29.54 \\ 
        \midrule
         AT \citep{madry2017towards} & 32-10 & P\&C &  $100^\dagger$ &   \no &   \no &  \no &      60.13 & 	24.76 \\
         TRADES \citep{zhang2019theoretically} & 32-10 & P\&C &  $100^\dagger$ &   \no &   \yes &  \no &      60.73 & 	24.90 \\
         MART \citep{wang2020improving} & 32-10 & P\&C &  $100^\dagger$ &   - &   - &  - &      54.08 & 	25.30 \\
         FAT \citep{zhang2020attacks} & 32-10 & P\&C &  $100^\dagger$ &   - &   - &  - &      \textbf{66.74} & 	20.88 \\
         AWP \citep{wu2020adversarial} & 32-10 & P\&C &  $100^\dagger$ &   - &   - &  - &      55.16 & 	25.16 \\
         GAIRAT \citep{zhang2020geometry} & 32-10 & P\&C &  $100^\dagger$ &   - &   - &  - &      58.43 & 	17.54 \\
         MAIL-AT \citep{liu2021probabilistic} & 32-10 & P\&C &  $100^\dagger$ &   - &   - &  - &      60.74 & 	22.44 \\
         MAIL-TR \citep{liu2021probabilistic} & 32-10 & P\&C &  $100^\dagger$ &   - &   \yes &  - &      60.13 & 	24.80 \\
         % $\lambda_{wd}=0.00281$* & 32-10 & P\&C &  200 &   \yes &   \no &  \no &      62.93 & 	26.82 \\
         $\bm{\lambda_{awd}} = 0.022$ + ASAM & 32-10 & P\&C &  100 &   \yes &   \no &  \no &      64.49 & 	\textbf{29.70} \\
         % $\bm{\lambda_{awd}} = 0.022$ & 32-10 & P\&C &  200 &   \yes &   \no &  \no &      64.37 & 	\textbf{29.55} \\
          
		\bottomrule
    \end{tabular}
	\caption{CIFAR-100 adversarial robustness performance of various strong methods.
	Adaptive weight decay with ASAM optimizer outperforms many strong baselines. 
	For experiments marked with * we do another round of hyper-parameter search. $\lambda_{AdaDecay}$ indicates using the work from \cite{nakamura2019adaptive}. 
	The columns represent the method, depth and width of the WideResNets used, augmentation, number of epochs, whether ASAM, TRADES \citep{zhang2019theoretically}, and Stochastic Weight Averaging \citep{izmailov2018averaging}, 
	were used in the training, followed by the natural accuracy and adversarial accuracy using AutoAttack.
	In the augmentation column, P\&C 
	is short for Pad and Crop. 
    The experiments with $\dagger$ are based on results from \citep{liu2021probabilistic} which use a custom choice of parameters to alleviate robust overfitting.  
    We also experimented with methods related to AWD such as LARS. We observed no improvement, so we do not report the results here. More details can be found in Appendix \ref{app:lars}. 
    } 
	\label{tab:sota}
\end{table*}

% \footnote{ }

Adaptive Weight Decay (AWD) can help improve the robustness over traditional weight decay on many datasets as summarized before in Table~\ref{tab:nat_perf_adv_trained}. 
In Table~\ref{tab:sota} we demonstrated that AWD when combined with advanced optimization methods such as ASAM can result in models which have good natural and robust accuracies when compared with advanced methods on the CIFAR-100 dataset. 
Table~\ref{tab:sotac10} compares AWD+ASAM with various advanced methods on the CIFAR-10 dataset. 
The hyper-parameters used in this experiment are similar to those used before for the CIFAR-100 dataset. 
As it can be seen, despite it's simplicity, AWD depicts improvements over very strong baselines on two extensively studied datasets of the adversarial machine learning domain \footnote{ImageNet robustness results can be seen in Appendix~\ref{app:imagenet}}. 

\begin{table*}[h]
    \hspace*{-0.5em}
	\centering
	\begin{tabular}{rccccccc}
    	\toprule
    	\multirow{2}{*}{Method} & \multirow{2}{*}{WRN} & \multirow{2}{*}{Aug} & \multirow{2}{*}{Epo} &  \multicolumn{2}{c}{CIFAR-10}\\ 
         & & & & Nat & AA \\ 
        \midrule
         AT \citep{madry2017towards} & 32-10 & P\&C &  $100^\dagger$ &        87.80 & 48.46\\
         TRADES \citep{zhang2019theoretically} & 32-10 & P\&C &  $100^\dagger$ &    86.36 & 53.40 \\
         MART \citep{wang2020improving} & 32-10 & P\&C &  $100^\dagger$ &        84.76 & 51.40  \\
         FAT \citep{zhang2020attacks} & 32-10 & P\&C &  $100^\dagger$ &      \textbf{89.70} & 47.48  \\
         AWP \citep{wu2020adversarial} & 32-10 & P\&C &  $100^\dagger$ &         57.55 & 53.08  \\
         GAIRAT \citep{zhang2020geometry} & 32-10 & P\&C &  $100^\dagger$ &      86.30 & 40.30  \\
         MAIL-AT \citep{liu2021probabilistic} & 32-10 & P\&C &  $100^\dagger$ &    84.83 & 47.10  \\
         MAIL-TR \citep{liu2021probabilistic} & 32-10 & P\&C &  $100^\dagger$ &    84.00 & 53.90  \\
         % $\lambda_{wd}=0.00281$* & 32-10 & P\&C &  200 &   \yes &   62.93 & 	26.82 \\
         $\bm{\lambda_{awd}} = 0.022$ + ASAM & 32-10 & P\&C &  100  & 88.55 & \textbf{54.04} \\
		\bottomrule
    \end{tabular}
	\caption{WRN32-10 models CIFAR-10 models trained with AWD when using the ASAM optimizer are more robust than models trained with various sophisticated algorithms from the literature.} 
	\label{tab:sotac10}
\end{table*}

\section{Additional Properties of Adaptive Weight Decay} \label{sec:properties}
Due to the properties mentioned before, such as reducing overfitting and resulting in networks with smaller weight-norms, Adaptive Weight Decay (AWD) can be seen as a good choice for other applications which can benefit from robustness. 
In particular, below in \ref{subsec:noisy} we study the effect of adaptive weight decay in the noisy label setting. 
More specifically, we show roughly 4\% accuracy improvement on CIFAR-100 and 2\% on CIFAR-10 for training on the 20\% symmetry label flipping setting \citep{bartlett2006convexity}. 
In addition, in the Appendix, we show the potential of AWD for reducing sensitivity to sub-optimal learning rates. 
Also, we show that networks which are naturally trained achieving roughly similar accuracy, once trained with adaptive weight decay, tend to have lower weight-norms. 
This phenomenon can have exciting implications for pruning networks \citep{lecun1989optimal,hassibi1992second}.

\subsection{Robustness to Noisy Labels} \label{subsec:noisy} % [#]
Popular vision datasets, such as MNIST \citep{lecunmnisthandwrittendigit2010}, CIFAR \citep{krizhevsky2009learning}, and ImageNet \citep{deng2009imagenet}, contain some amount of label noise \citep{yun2021re, zhang2017method}. 
While some studies provide methods for identifying and correcting such errors \citep{yun2021re, muller2019identifying, al2018labeling, kuriyama2020autocleansing}, others provide training algorithms that avoid over-fitting, or even better, avoid fitting the incorrectly labeled examples entirely \citep{jiang2018mentornet, song2019selfie, jiang2020beyond}.

\begin{figure*}[ht]
    \vspace{-.75em}
    \centering
    \subfigure[]{\includegraphics[width=0.45\columnwidth]{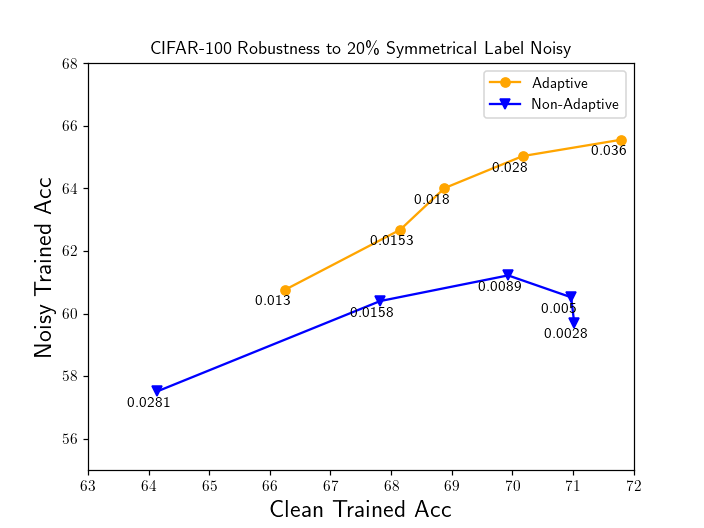}\label{fig:noisy_c100_20}}
    \subfigure[]{\includegraphics[width=0.45\columnwidth]{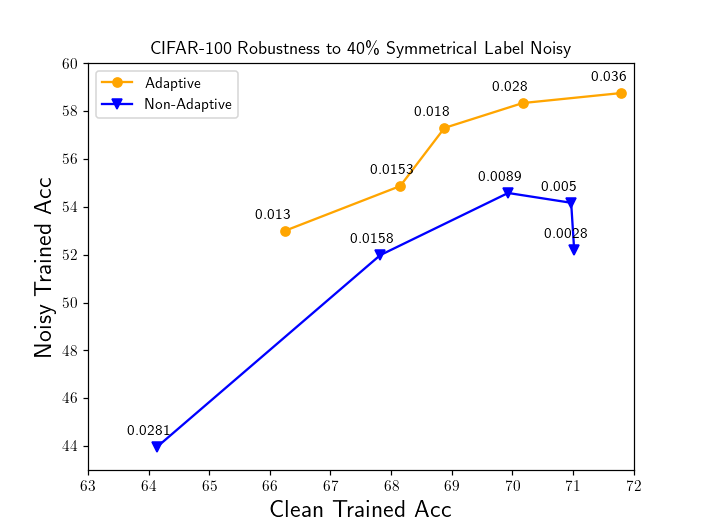}\label{fig:noisy_c100_40}}
    \subfigure[]{\includegraphics[width=0.45\columnwidth]{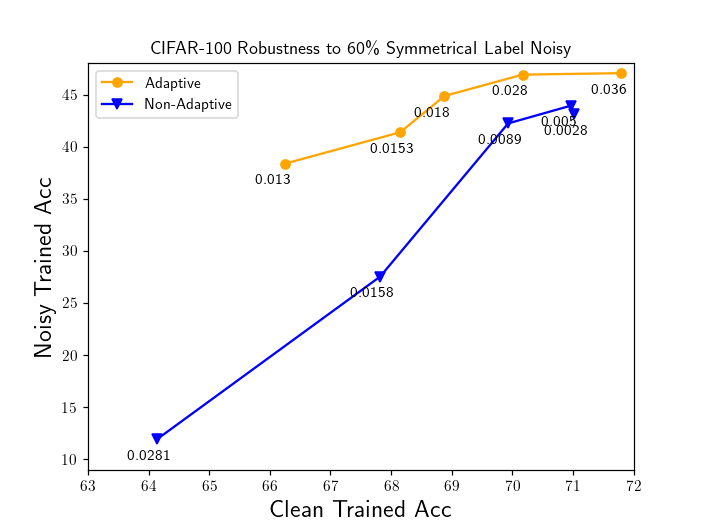}\label{fig:noisy_c100_60}}
    \subfigure[]{\includegraphics[width=0.45\columnwidth]{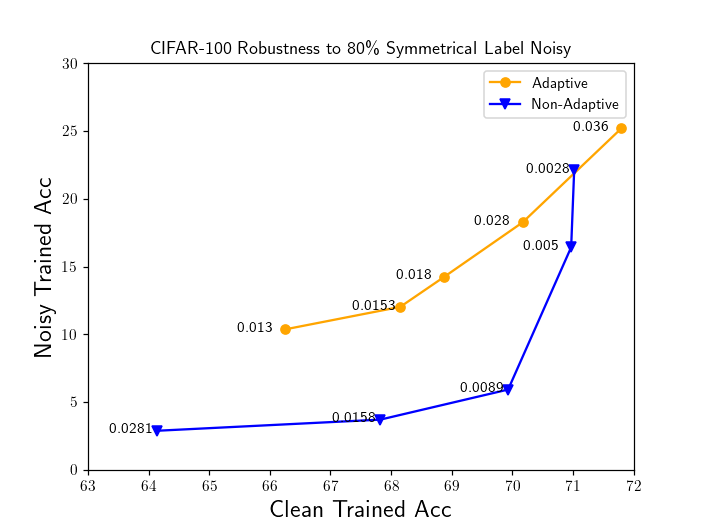}\label{fig:noisy_c100_80}}
    \caption{Comparison of similarly performing networks once trained on CIFAR-100 clean data, after training on 20\% (a), 40\% (b), 60\% (c), and 80\% (d).
    Networks trained with adaptive weight decay are less sensitive to label noise compared to ones trained with non-adaptive weight decay.}
    \label{fig:noisy_c100}
    % \vspace{-3.em}
\end{figure*}

In this section, we perform a preliminary investigation of adaptive weight decay's resistance to fitting training data with label noise. 
Following previous studies, we use symmetry label flipping \citep{bartlett2006convexity} to create noisy data for CIFAR-10 and CIFAR-100 and use ResNet34 as the backbone. 
Other experimental setup details can be found in Appendix~\ref{app:setup_noisy}. 
Similar to the previous section, we test different hyper-parameters for adaptive and non-adaptive weight decay. 
To ease comparison in this setting, we train two networks for each hyper-parameter: 1- with a certain degree of label noise and 2- with no noisy labels. 
We then report the accuracy on the clean label test set. 
The test accuracy on the second network -- one which is trained with no label noise -- is just the clean accuracy.  
Having the clean accuracy coupled with the accuracy after training on noisy data enables an easier understanding of the sensitivity of each training algorithm and choice of hyper-parameter to label noise. 
Figure~\ref{fig:noisy_c100} gathers the results of the noisy data experiment on CIFAR-100.

% Figure~\ref{fig:noisy_c100} illustrates that networks trained with adaptive weight decay have a smaller drop in performance when there is label noise in the training set. 
Figure~\ref{fig:noisy_c100} demonstrates that networks trained with adaptive weight decay exhibit a more modest decline in performance when label noise is present in the training set.
For instance, Figure~\ref{fig:noisy_c100_20} shows that ${\lambda_{awd}}=0.028$ for adaptive and $\lambda_{wd}=0.0089$ for non-adaptive weight decay achieve roughly 70\% accuracy when trained on clean data, while the adaptive version achieves 4\% higher accuracy when trained on the noisy data. 
Appendix~\ref{app:c10noisy} includes similar results for CIFAR-10. 

% Intuitively, from eq.~\ref{eq:awd}, towards the end of the training, where the examples with label noise are producing large gradients, adaptive weight decay increases the penalty for weight decay, which prevents fitting the noisy data by regularizing the gradients. 

Intuitively, based on eq.~\ref{eq:awd}, in the later stages of training, when examples with label noise generate large gradients, adaptive weight decay intensifies the penalty for weight decay. This mechanism effectively prevents the model from fitting the noisy data by regularizing the gradients.

\section{Conclusion}
Regularization methods for a long time have aided deep neural networks in generalizing on data not seen during training. 
Due to their significant effects on the outcome, it is crucial to have the right amount of regularization and correctly tune training hyper-parameters. 
We propose Adaptive Weight Decay (AWD), which is a simple modification to weight decay – one of the most commonly employed regularization methods.
In our study, we conduct a comprehensive comparison between AWD and non-adaptive weight decay in various settings, including adversarial robustness and training with noisy labels. Through rigorous experimentation, we demonstrate that AWD consistently yields enhanced robustness.
By conducting experiments on diverse datasets and architectures, we provide empirical evidence to showcase the effectiveness of our approach in mitigating robust overfitting.

\section{Acknowledgement}
We extend our sincere appreciation to Zhile Ren for their invaluable support and perceptive contributions during the publication of this manuscript.

\clearpage

\bibliography{iclr2023_conference}
\bibliographystyle{icml2023}

%%%%%%%%%%%%%%%%%%%%%%%%%%%%%%%%%%%%%%%%%%%%%%%%%%%%%%%%%%%%%%%%%%%%%%%%%%%%%%%
%%%%%%%%%%%%%%%%%%%%%%%%%%%%%%%%%%%%%%%%%%%%%%%%%%%%%%%%%%%%%%%%%%%%%%%%%%%%%%%
% APPENDIX
%%%%%%%%%%%%%%%%%%%%%%%%%%%%%%%%%%%%%%%%%%%%%%%%%%%%%%%%%%%%%%%%%%%%%%%%%%%%%%%
%%%%%%%%%%%%%%%%%%%%%%%%%%%%%%%%%%%%%%%%%%%%%%%%%%%%%%%%%%%%%%%%%%%%%%%%%%%%%%%
\newpage
\appendix
\onecolumn

\section{Experimental Setup} \label{app:setup}
In this section, we include the experimental setup used to produce the experiments throughout this paper. 
We include all hyperparameters used for all experiments, unless explicitly mentioned otherwise. 
For all experiments, we use SGD optimizers with momentum $\mu=0.9$. 
We use a Cosine learning rate schedule with no warm-up and with the value of $0$ for the final value. 
The weight decay and learning rate for experiments that have not been clearly specified are $\lambda_{wd}=0.0005$ and $lr=0.1$.
We train all networks for $200$ epochs with a batch-size of $128$.
For all experiments that use ASAM, we use the hyper-parameters the original paper suggests. 

\subsection{Adversarial Training} \label{app:setup_adv}
We use a pre-activation WideResNet28 with a width of $10$. 
We use $\ell_{\infty}=8/255$ PGD attack with the step size of $2/255$. 
We use 7 steps for creation of adversarial examples for training and use the minimum accuracy produced by AutoAttack for the test. 
We keep $10\%$ of training data for validation and use it to do early stopping.
For $\lambda_{wd}$ of non-adaptive weight decay, we fit a geometric sequence of length $17$ starting from $0.00005$ and ending at $0.005$. %
For the adaptive weight decay hyper parameter (${\lambda_{awd}}$), we fit a geometric sequence of length $9$ starting from $0.02$ and ending at $0.04$. 

\subsection{Noisy Label Training} \label{app:setup_noisy}
We use a ResNet34 as our architecture. 
For each setting we used 5 different hyperparameters for adaptive and non-adaptive weight decays. 
For CIFAR-10, we use $\lambda_{wd} \in \{ 0.0028, 0.005, 0.0089, 0.0158,  0.0281 \}$ and ${\lambda_{awd}} \in \{ 0.036, 0.028, 0.018, 0.0153, 0.018 \}$ 
and for CIFAR-100, we use $\lambda_{wd} \in \{ 0.0281, 9.9158, 0.0089, 0.005, 0.0028 \}$ and ${\lambda_{awd}} \in \{ 0.013, 0.0153, 0.018, 0.028, 0.036 \}$.

\section{Implementation} \label{app:impl}
In this section, we discuss the details of implementation of adaptive weight decay. 
The method is not really susceptible to the exact implementation details discussed here, however, to be perfectly candid, we include all details here. 
First, let us assume that we desire to implement adaptive weight decay using ${\lambda_{awd}}=0.016$ as the hyperparameter. 
We know that $ \lambda_{t} = \frac{\| \nabla w_t \| 0.016}{\| w_t \|}$.
Please note that $\| \nabla w_t \|$ requires knowing the gradients of the loss w.r.t. the network's parameters. 
Meaning that to compute the $\lambda_{t}$ for every step, we have to call a backward pass on the actual parameters of the network. 
After this step, given the fact that we know both $\| w_t \|$ and $\| \nabla w_t \|$, we can compute $\lambda_{t}$.

\section{Extra Results}
Here, we provide the extra results and figures not included in the body of the paper.

\subsection{Varying the architecture} \label{app:arch}
To measure the dependency of AWD to architecture choices, we perform similar set of experiments to that in the main body but by varying the architecture. 
In particular, Table~\ref{tab:arch} summarizes the best performing hyper-parameter for Adaptive and Non-Adaptive Weight Decay methods on CIFAR-100 for various architectures. 
As it can be seen, AWD's considerable boost in robustness is not dependent on architecture. 
Not only adaptive weight decay outperforms the non-adaptive weight decay trained model in terms of robust accuracy, it is also superior in terms of the natural accuracy. 
Models trained with AWD have considerably smaller weight-norms to those trained with non-adaptive weight decay. 
In the last column, we report the value of the total loss of the non-adaptive weight decay for the best tuned non-adaptive ($\lambda_{wd}$) for that dataset which is found by doing a grid search. 
Interestingly, when we measure the non-adaptive total loss (eq.~\ref{eq:adv_training}) on the training set, we observe that networks trained with the adaptive method often have smaller non-adaptive training loss even though in AWD we have not optimized that loss directly.
\begin{table*}[h]
    % \vspace{-1.em}
	\centering
	\begin{tabular}{rccccccc}
    	\toprule
    	Method &						Arch &	Opt & 	$\|W\|_2$ & 	Nat Acc &		AutoAtt & $Xent + \frac{\lambda_{wd}^* \cdot \|W\|_2^2}{2}$ \\     	

    	\midrule
        $\lambda_{wd} = 0.00211$ 	&	\multirow{2}{*}{ResNet18} &	SGD & 			22.69 & 		58.04 &  		21.94 & 2.43 \\ 
    	${\lambda_{awd}}=0.02181$ &		&  SGD & 			\textbf{12.89} & 		\textbf{58.46} &			\textbf{24.98} & \textbf{2.32} \\ 
        \midrule
        $\lambda_{wd} = 0.00211$ 	&	\multirow{2}{*}{ResNet50} &	SGD & 			24.50 & 		59.06 &  		22.30 & 2.27 \\ 
    	${\lambda_{awd}}=0.02$ &		&  SGD & 			\textbf{13.87} & 		\textbf{60.60} &			\textbf{26.73} & \textbf{2.02} \\ 
    \midrule
     $\lambda_{wd} = 0.0005$ 	&	\multirow{2}{*}{WRN28-10} &	SGD & 			51.32 & 		60.15 &  		22.53 & \textbf{0.67} \\ 
    	${\lambda_{awd}}=0.022$ &							&	SGD & 			\textbf{13.41} & 		\textbf{61.39} &			\textbf{27.15} & 1.51 \\   
        
		\bottomrule
    \end{tabular}
	\caption{Adversarial robustness of PGD-7 adversarially trained CIFAR-100 networks using adaptive and non-adaptive weight decay on different architectures.  }% \ali{updated and new section.}}
	\label{tab:arch}
\end{table*}

\subsection{Varying the Attack Budget } \label{app:various_eps}
As discussed in Section \ref{subsec:results_robustness}, adaptive weight decay improves the performance of the network both in terms of robustness accuracy, as well as the natural accuracy, when the attack budget $\epsilon = 8$. 
To further show the applicability of the AWD method, we reproduced the CIFAR-10 and CIFAR-100 experiments with various $\epsilon$ budgets. 
We use the same budget $\epsilon$ for both training and evaluation of the network. 
Table \ref{tab:various_eps} summarizes these results. 

\begin{table*}[t]
	\centering
    \hspace*{-6em}
	\begin{tabular}{rccccccccccccc}
    	\toprule
    	$\epsilon$ & Data & Method & Nat & 20 & 40 & 60 & 80 & 100 & AA-SQ & AA-CE & AA-FAB & AA-T & AA\\
        \midrule

        2 & C10 & $\lambda_{wd} = $0.00089 & 94.2 & 83.2 & 83.1 & 83.2 & 83.1 & 83.1 & 86.9 & 82.7 & 82.7 & 82.5 & 82.5\\
        2 & C10 & $\lambda_{awd} = $0.02181 & \textbf{94.3} & \textbf{83.6} & \textbf{83.6} & \textbf{83.6} & \textbf{83.6} & \textbf{83.6} & \textbf{87} & \textbf{83.2} & \textbf{83.1} & \textbf{83} & \textbf{83}\\
        2 & C100 & $\lambda_{wd} = $0.00067 & 74.8 & 55.8 & 55.8 & 55.7 & 55.7 & 55.7 & 59.2 & 54.8 & 52.9 & 52.7 & 52.7\\
        2 & C100 & $\lambda_{awd} = $0.02181 & \textbf{75.2} & \textbf{56.7} & \textbf{56.7} & \textbf{56.7} & \textbf{56.6} & \textbf{56.7} & \textbf{59.6} & \textbf{56} & \textbf{53.7} & \textbf{53.4} & \textbf{53.4}\\
        
        \midrule
        
        4 & C10 & $\lambda_{wd} = $0.00158 & 91.7 & 70.7 & 70.7 & 70.5 & 70.6 & 70.6 & 75.3 & 69.2 & 69.4 & 69 & 69\\
        4 & C10 & $\lambda_{awd} = $0.02181 & \textbf{92} & \textbf{73.1} & \textbf{73.1} & \textbf{73.1} & \textbf{73} & \textbf{73} & \textbf{77.8} & \textbf{72} & \textbf{71.7} & \textbf{71.3} & \textbf{71.3}\\
        4 & C100 & $\lambda_{wd} = $0.00089 & 69.4 & 42 & 42 & 42 & 41.9 & 41.9 & 44.8 & 40.4 & 38.9 & 38.6 & 38.6\\
        4 & C100 & $\lambda_{awd} = $0.02181 & \textbf{71.5} & \textbf{46.8} & \textbf{46.7} & \textbf{46.7} & \textbf{46.7} & \textbf{46.7} & \textbf{48.3} & \textbf{45.2} & \textbf{41.2} & \textbf{40.8} & \textbf{40.8}\\
        
        \midrule
        
        6 & C10 & $\lambda_{wd} = $0.00119 & 88.7 & 59 & 58.9 & 58.9 & 58.9 & 58.9 & 63.9 & 56 & 56.5 & 55.8 & 55.8\\
        6 & C10 & $\lambda_{awd} = $0.02181 & \textbf{90} & \textbf{62.4} & \textbf{62.3} & \textbf{62.4} & \textbf{62.3} & \textbf{62.3} & \textbf{67.3} & \textbf{60.3} & \textbf{60} & \textbf{59.5} & \textbf{59.5}\\
        6 & C100 & $\lambda_{wd} = $0.00067 & 64.7 & 32.8 & 32.6 & 32.7 & 32.6 & 32.6 & 35 & 30.9 & 29.5 & 29.2 & 29.2\\
        6 & C100 & $\lambda_{awd} = $0.02181 & \textbf{66.3} & \textbf{39.6} & \textbf{39.5} & \textbf{39.5} & \textbf{39.5} & \textbf{39.5} & \textbf{39.8} & \textbf{37.7} & \textbf{33.4} & \textbf{33.1} & \textbf{33.1}\\
        
        \midrule
        
        8 & C10 & $\lambda_{wd} = $0.00158 & 84 & 49.5 & 49.2 & 49.4 & 49.4 & 49.4 & 54.3 & 46.5 & 45.1 & 44.7 & 44.7\\
        8 & C10 & $\lambda_{awd} = $0.02181 & \textbf{87.3} & \textbf{53.9} & \textbf{53.8} & \textbf{53.7} & \textbf{53.8} & \textbf{53.8} & \textbf{58.1} & \textbf{51.4} & \textbf{50.1} & \textbf{49.6} & \textbf{49.6}\\
        8 & C100 & $\lambda_{wd} = $0.00158 & 56.5 & 27.7 & 27.7 & 27.7 & 27.5 & 27.5 & 28.9 & 25.9 & 22.6 & 22.4 & 22.4\\
        8 & C100 & $\lambda_{awd} = $0.02181 & \textbf{61.6} & \textbf{33.1} & \textbf{33} & \textbf{33.1} & \textbf{33.1} & \textbf{33} & \textbf{32.5} & \textbf{31} & \textbf{26.7} & \textbf{26.4} & \textbf{26.4}\\
        
        \midrule
        
        16 & C10 & $\lambda_{wd} = $0.00119 & 70.4 & 32 & 31.7 & 31.6 & 31.5 & 31.7 & 30.5 & 27.1 & 22.6 & 21.6 & 21.6\\
        16 & C10 & $\lambda_{awd} = $0.02181 & \textbf{71.9} & \textbf{34} & \textbf{33.8} & \textbf{33.7} & \textbf{33.7} & \textbf{33.7} & \textbf{33.2} & \textbf{29.6} & \textbf{26.1} & \textbf{25.3} & \textbf{25.3}\\
        16 & C100 & $\lambda_{wd} = $0.00281 & 38.3 & 16.5 & 16.5 & 16.5 & 16.5 & 16.5 & 14.2 & 14.8 & 11.3 & 11 & 11\\
        16 & C100 & $\lambda_{awd} = $0.02181 & \textbf{41.5} & \textbf{19.8} & \textbf{19.7} & \textbf{19.6} & \textbf{19.5} & \textbf{19.5} & \textbf{17} & \textbf{17.7} & \textbf{13.9} & \textbf{13.4} & \textbf{13.4}\\
		\bottomrule
    \end{tabular}
	\caption{Adversarial robustness of PGD-7 adversarially trained WRN28-10 networks using adaptive and non-adaptive weight decay under various choices of $\epsilon$. Table summarizes the best performing hyper-parameter for non adaptive method, compared with the fixed hyper-parameter $\lambda_{awd}=0.02181$. Not only the adaptive method outperforms the non-adaptive method in terms of robust accuracy, it is also superior in terms of the natural accuracy. Columns with numbers in the header, show the resulting robustness performance evaluated using multi-step PGD attacks. The final robust accuracy, which is the minimum accuracy over all attacks is gathered in the last column (AA.)}
	\label{tab:various_eps}
\end{table*}

As it can be seen, regardless of the attack budget (i.e. $\epsilon$), the AWD trained models always outperform the non-adaptive counter parts, both in terms of natural and robustness accuracy.

\subsection{Varying the number of Epochs}\label{app:epoch_vary}
In this section, we investigate AWD's performance and sensitivity to the length of training (number of epochs). To do so, we adversarially train WRN28-10 networks using PGD-7 with $\epsilon=8$. 
Table \ref{tab:varying_epochs} summarizes these results for the CIFAR-10 and CIFAR-100 datasets. As it can be seen, for various choices of training epochs, AWD outperfroms traditional weight decay both in natural accuracy and in robustness accuracy measured with AutoAttack (AA).

\begin{table*}[t]
	\centering
    \hspace*{-6.5em}
	\begin{tabular}{rccccccccccccc}
    	\toprule
    	Epoch & Data & Method & Nat & 20 & 40 & 60 & 80 & 100 & AA-SQ & AA-CE & AA-FAB & AA-T & AA\\

        \midrule

        50 & C10 & $\lambda_{wd}=$0.00158 & 86.5 & 52.6 & 52.4 & 52.4 & 52.4 & 52.5 & 56.7 & 49.5 & 48.5 & 48.0 & 48.0\\
        50 & C10 & $\lambda_{awd}=$0.02181 & \textbf{87.1} & \textbf{54.3} & \textbf{54.0} & \textbf{54.0} & \textbf{54.0} & \textbf{54.1} & \textbf{58.7} & \textbf{51.3} & \textbf{50.0} & \textbf{49.5} & \textbf{49.5}\\
        50 & C100 & $\lambda_{wd}=$0.00211 & 59.5 & 29.8 & 29.7 & 29.7 & 29.7 & 29.6 & 31.1 & 28.0 & 25.3 & 25.1 & 25.1\\
        50 & C100 & $\lambda_{awd}=$0.02181 & \textbf{61.9} & \textbf{32.4} & \textbf{32.4} & \textbf{32.4} & \textbf{32.4} & \textbf{32.3} & \textbf{32.8} & \textbf{30.5} & \textbf{26.7} & \textbf{26.4} & \textbf{26.4}\\
        
        \midrule
        
        100 & C10 & $\lambda_{wd}=$0.00211 & 85.8 & 50.8 & 50.6 & 50.5 & 50.6 & 50.6 & 55.4 & 47.7 & 47.1 & 46.6 & 46.6\\
        100 & C10 & $\lambda_{awd}=$0.02181 & \textbf{87.7} & \textbf{55.1} & \textbf{55.0} & \textbf{55.0} & \textbf{54.9} & \textbf{54.9} & \textbf{59.6} & \textbf{52.5} & \textbf{51.5} & \textbf{51.2} & \textbf{51.2}\\
        100 & C100 & $\lambda_{wd}=$0.00281 & 58.2 & 28.2 & 28.1 & 28.0 & 28.1 & 28.1 & 29.4 & 26.3 & 23.7 & 23.4 & 23.4\\
        100 & C100 & $\lambda_{awd}=$0.02181 & \textbf{62.5} & \textbf{33.3} & \textbf{33.3} & \textbf{33.3} & \textbf{33.2} & \textbf{33.2} & \textbf{33.0} & \textbf{31.2} & \textbf{27.1} & \textbf{26.7} & \textbf{26.7}\\
        
        \midrule
        
        200 & C10 & $\lambda_{wd}=$0.00158 & 84.1 & 50.3 & 50.1 & 50.1 & 50.0 & 50.0 & 54.8 & 47.4 & 46.2 & 45.7 & 45.7\\
        200 & C10 & $\lambda_{awd}=$0.02181 & \textbf{87.3} & \textbf{54.2} & \textbf{54.0} & \textbf{54.0} & \textbf{54.0} & \textbf{54.0} & \textbf{58.5} & \textbf{51.4} & \textbf{50.5} & \textbf{50.0} & \textbf{50.0}\\
        200 & C100 & $\lambda_{wd}=$0.0005 & 60.5 & 25.2 & 25.2 & 25.1 & 25.0 & 25.1 & 27.0 & 23.1 & 22.4 & 22.2 & 22.2\\
        200 & C100 & $\lambda_{awd}=$0.02181 & \textbf{62.0} & \textbf{33.1} & \textbf{32.9} & \textbf{33.0} & \textbf{32.9} & \textbf{32.9} & \textbf{32.1} & \textbf{30.9} & \textbf{26.8} & \textbf{26.4} & \textbf{26.4}\\
        
        \midrule
        
        300 & C10 & $\lambda_{wd}=$0.00089 & 86.2 & 48.9 & 48.8 & 48.6 & 48.6 & 48.6 & 52.8 & 44.9 & 45.2 & 44.5 & 44.5\\
        300 & C10 & $\lambda_{awd}=$0.02181 & \textbf{87.3} & \textbf{52.8} & \textbf{52.7} & \textbf{52.8} & \textbf{52.8} & \textbf{52.7} & \textbf{57.4} & \textbf{50.3} & \textbf{49.4} & \textbf{48.8} & \textbf{48.8}\\
        300 & C100 & $\lambda_{wd}=$0.00028 & 59.5 & 25.6 & 25.5 & 25.5 & 25.5 & 25.5 & 27.0 & 23.6 & 22.6 & 22.3 & 22.3\\
        300 & C100 & $\lambda_{awd}=$0.02181 & \textbf{62.0} & \textbf{33.1} & \textbf{32.9} & \textbf{33.0} & \textbf{32.9} & \textbf{32.9} & \textbf{32.1} & \textbf{30.9} & \textbf{26.8} & \textbf{26.4} & \textbf{26.4}\\

		\bottomrule
    \end{tabular}
	\caption{Adversarial robustness of PGD-7 adversarially trained WRN28-10 networks using adaptive and non-adaptive weight decay. Table summarizes the best performing hyper-parameter for non adaptive method, compared with the fixed hyper-parameter $\lambda_{awd}=0.022$. Not only the adaptive method outperforms the non-adaptive method in terms of robust accuracy, it is also superior in terms of the natural accuracy. Columns with numbers in the header, show the resulting robustness performance evaluated using multi-step PGD attacks.}
	\label{tab:varying_epochs}
\end{table*}

To further study the performance AWD in low epoch settings, we also reproduced the results of Table \ref{tab:varying_epochs} with a WRN32-10 architecture and by varying the number of epochs. 
Table \ref{tab:very_low_epochs} summarizes these results. As illustrated, AWD's performance is not very sensitive to training time. While at 100 epochs, AWD's results (both nat and robustness) are comparable to models trained with 200 epochs, we found that even if we further reduce the epochs to 50, we do not see a big degradation of robust accuracy, although the natural accuracy degrades slightly. 

\begin{table*}[t]
	\centering
    % \hspace*{-6.5em}
	\begin{tabular}{rccccc}
    	\toprule
    	Epoch & Data & Method & Network & Nat & AA\\

        \midrule
        5 & C100 & $\lambda_{awd}=0.022$ + ASAM & WRN32-10 & 26.79 & 11.03 \\ 
        10 & C100 & $\lambda_{awd}=0.022$ + ASAM & WRN32-10 & 38.10 & 15.92 \\ 
        20 & C100 & $\lambda_{awd}=0.022$ + ASAM & WRN32-10 & 51.27 & 21.67 \\ 
        30 & C100 & $\lambda_{awd}=0.022$ + ASAM & WRN32-10 & 58.10 & 25.03 \\ 
        40 & C100 & $\lambda_{awd}=0.022$ + ASAM & WRN32-10 & 62.01 & 27.51 \\ 
        50 & C100 & $\lambda_{awd}=0.022$ + ASAM & WRN32-10 & 62.85 & 29.25 \\ 
        100 & C100 & $\lambda_{awd}=0.022$ + ASAM & WRN32-10 & \textbf{64.49} & 29.70 \\ 
        150 & C100 & $\lambda_{awd}=0.022$ + ASAM & WRN32-10 & 64.17 & \textbf{29.94} \\ 
        200 & C100 & $\lambda_{awd}=0.022$ + ASAM & WRN32-10 & 64.37 & 29.55 \\ 
        250 & C100 & $\lambda_{awd}=0.022$ + ASAM & WRN32-10 & 63.24 & 29.68 \\ 
        300 & C100 & $\lambda_{awd}=0.022$ + ASAM & WRN32-10 & 63.35 & 29.28 \\ 
        
		\bottomrule
    \end{tabular}
	\caption{Adversarial robustness of PGD-7 adversarially trained WRN32-10 networks using adaptive weight decay with fixed hyper-parameter $\lambda_{awd}=0.022$ and varying number of training epochs. The adaptive method has an acceptable performance in settings with low training epochs, even as low as 50 epochs. }
	\label{tab:very_low_epochs}
\end{table*}

\subsection{CIFAR-100 robustness and Adaptive Weight Decay} \label{app:c100_robust_overf}
\begin{figure}[h]
    \centering
    \subfigure[]{\includegraphics[width=0.32\columnwidth]{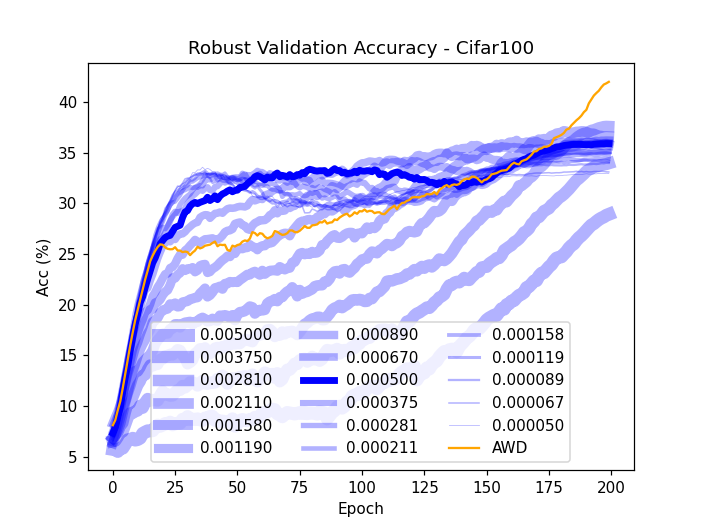}\label{fig:adv_val_acc_c100}}
    \subfigure[]{\includegraphics[width=0.32\columnwidth]{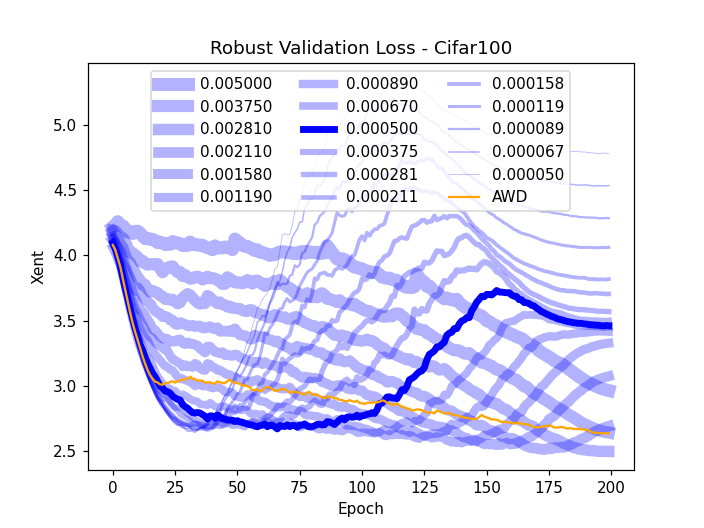}\label{fig:adv_val_loss_c100}}
    \subfigure[]{\includegraphics[width=0.32\columnwidth]{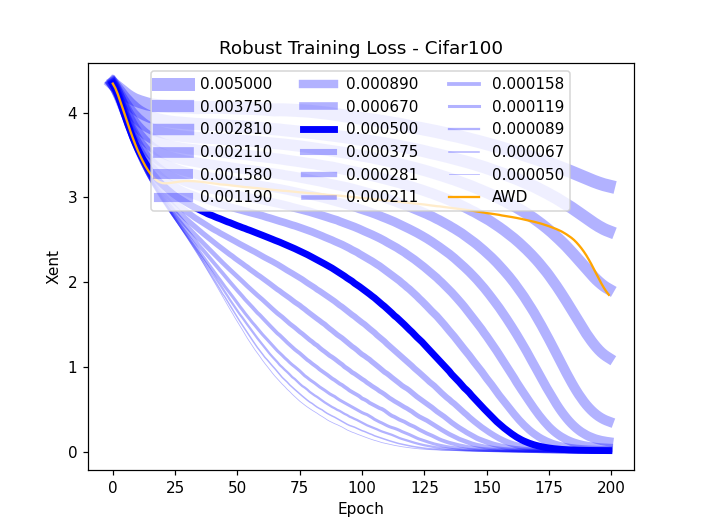}\label{fig:adv_train_loss_c100}}
    \caption{ Robust validation accuracy (a), validation loss (b), and training loss (c) on CIFAR-100 held-out validation subset. $\lambda_{wd} = 0.0005$ is the best performing hyper-parameter we found by doing a grid-search. The thickness of the plot-lines correspond to the magnitude of the weight-norm penalties. As it can be seen by (a) and (b), networks trained by small values of $\lambda_{wd}$ suffer from robust-overfitting, while networks trained with larger values of $\lambda_{wd}$ do not suffer from robust overfitting but the larger $\lambda_{wd}$ further prevents the network from fitting the data (c) resulting in reduced overall robustness. }
    \label{fig:adv_over_fitting_c100}
\end{figure}

Figure~\ref{fig:adv_over_fitting_c100} shows similar results to that of CIFAR-10 presented in Figure~\ref{fig:adv_over_fitting}. 

\subsection{CIFAR-10 robustness to noisy labels}\label{app:c10noisy}
\begin{figure}[h]
    \centering
    \subfigure[]{\includegraphics[width=0.49\columnwidth]{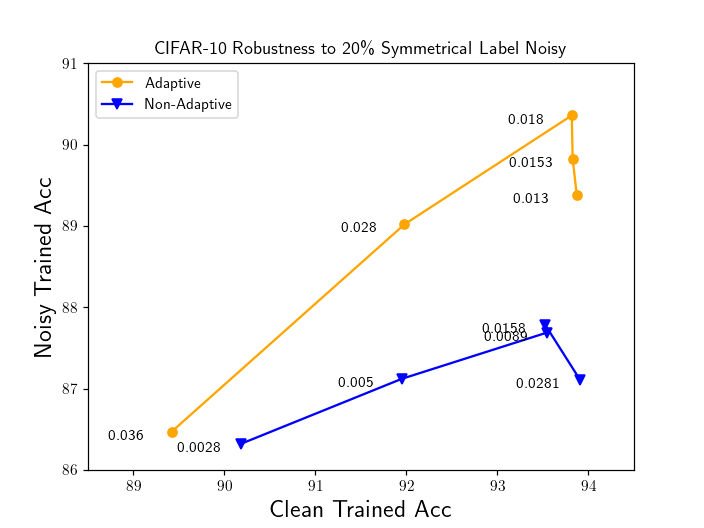}\label{fig:noisy_c10_20}}
    \subfigure[]{\includegraphics[width=0.49\columnwidth]{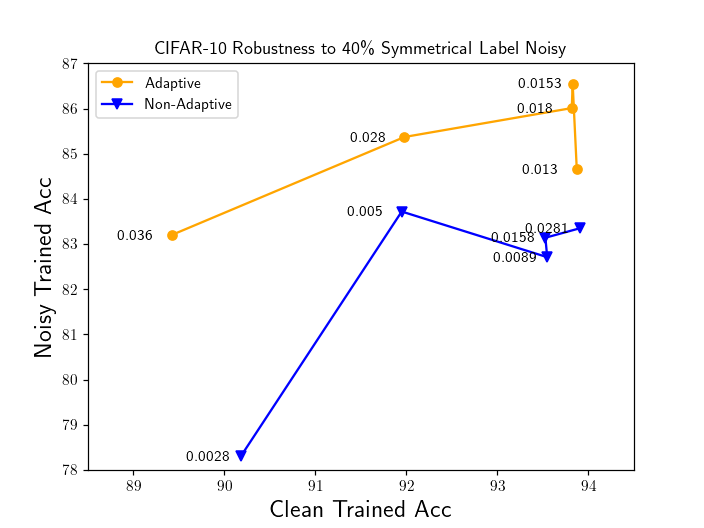}\label{fig:noisy_c10_40}}
    \subfigure[]{\includegraphics[width=0.49\columnwidth]{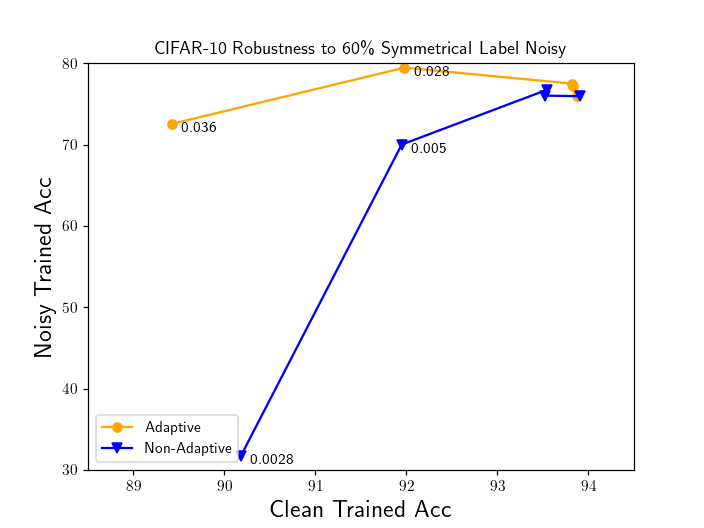}\label{fig:noisy_c10_60}}
    \subfigure[]{\includegraphics[width=0.49\columnwidth]{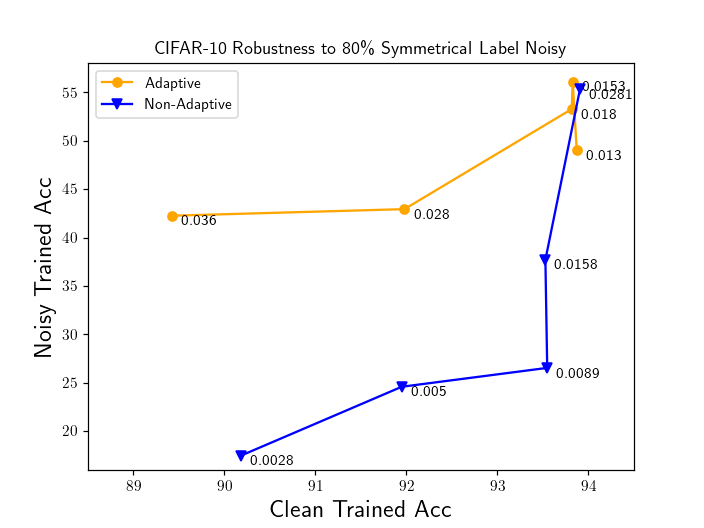}\label{fig:noisy_c10_80}}
    \caption{Comparison of similarly performing networks once trained on CIFAR-100 clean data, after training on 20\% (a), 40\% (b), 60\% (c), and 80\% (d) noisy data.
    Networks trained with adaptive weight decay outperform non-adaptive trained networks.}
    \label{fig:noisy_c10}
\end{figure}

The results of the experiments for training classifiers on the CIFAR-10 dataset with noisy labels can be seen in Figure~\ref{fig:noisy_c10} which yields similar conclusions to that of CIFAR-100. 

\subsection{2D Grid search for best parameter values for ResNet32} \label{sec:wd}
\begin{figure}[h]
    \centering
    \subfigure[]{\includegraphics[width=0.49\columnwidth]{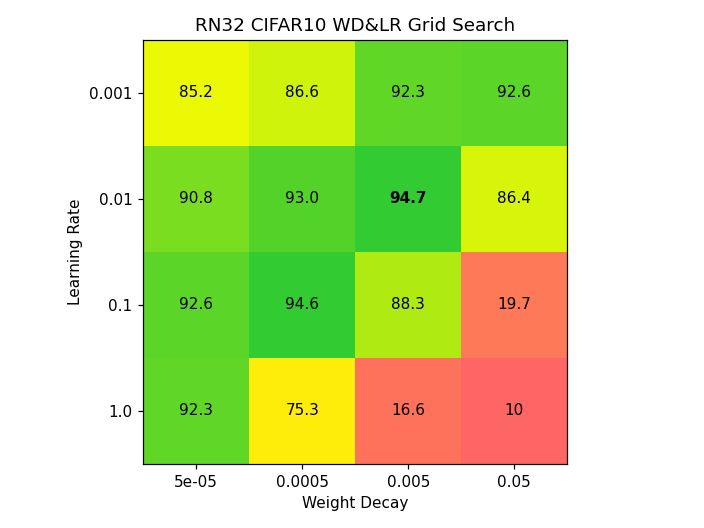}\label{fig:wd_rn32_c10}}
    \subfigure[]{\includegraphics[width=0.49\columnwidth]{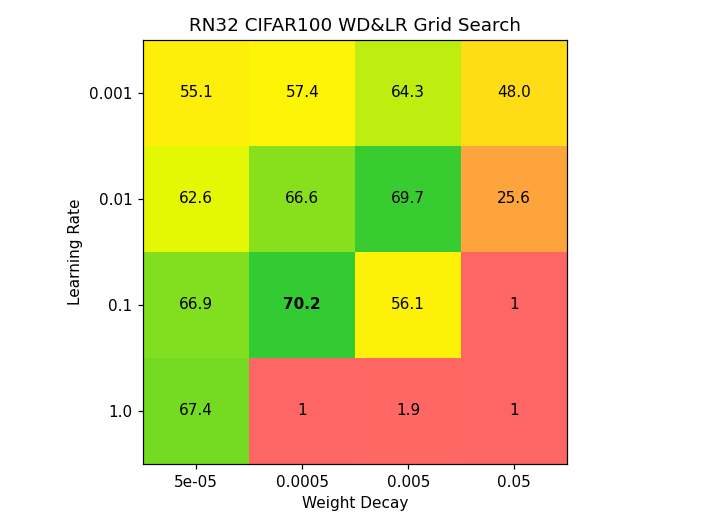}\label{fig:wd_rn32_c100}}
    \caption{Grid Search on different values of learning rate and weight decay on accuracy of ResNet32 on (a) CIFAR10 and (b) CIFAR100.}\label{fig:Resnet322DGridsearch}
\end{figure}

The importance of the 2D grid search on learning-rate and weight decay hyper-parameters are not network dependent. And we can see how these values are tied together for ResNet32 in Figure~\ref{fig:Resnet322DGridsearch}.

\subsection{Visualizing images from the CIFAR-100 training set where best AWD models do not fit.}\label{app:unfit_c100}
\begin{figure}[h]
    \centering

\includegraphics[width=1.0\columnwidth]{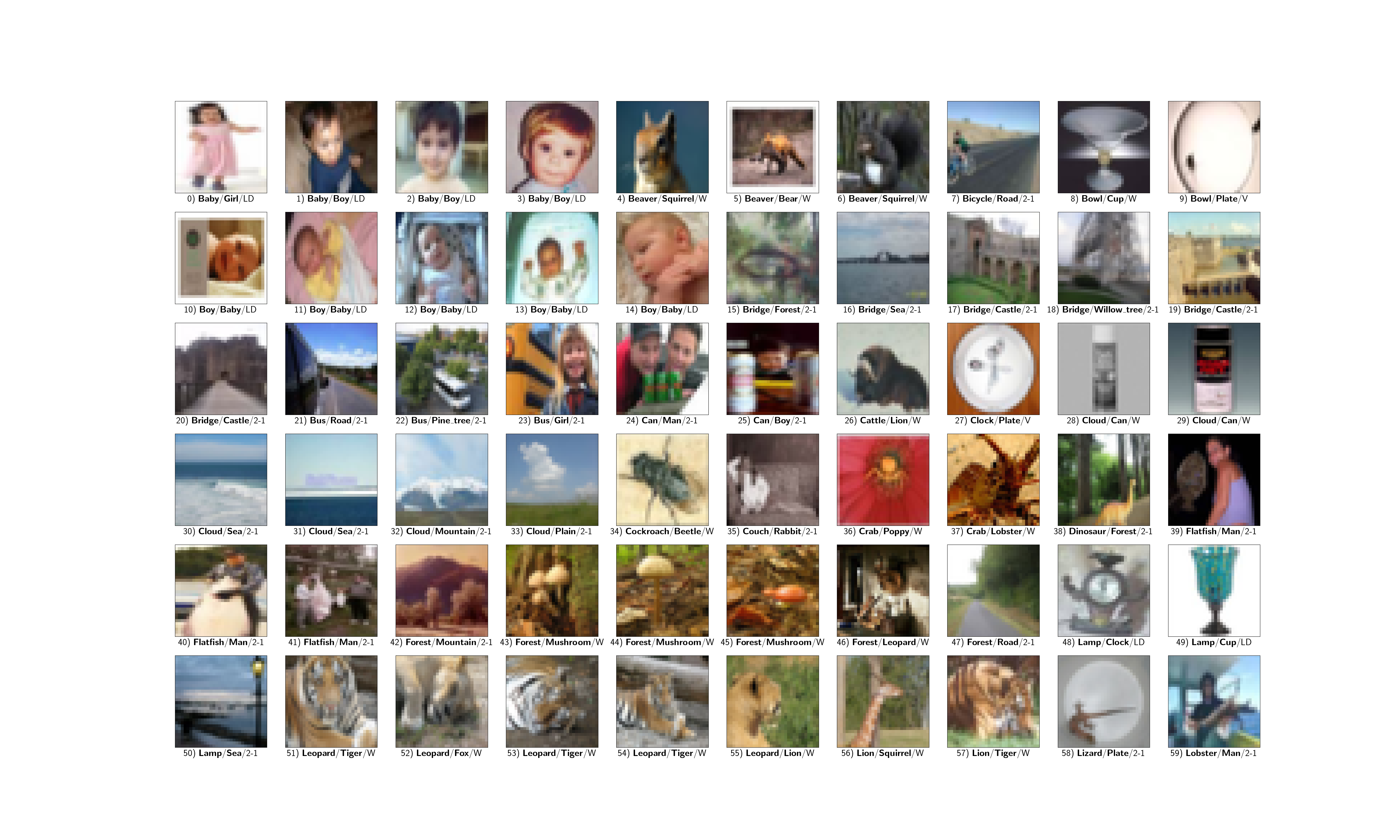}
    % \vspace{-5mm}
    \caption{Examples from CIFAR-100 training dataset that have noisy labels. For every image we state [dataset label/ prediction of classifier trained with AWD / our category of noisy case]. We classify these noisy labels into several categories: \textbf{W}: Wrong Labels where the picture is clear enough to comprehend the correct label.; \textbf{2-1}: Two Objects from CIFAR-100 in one image, but only one label is given in the dataset; 
    \textbf{LD}: Loosely Defined Classes where there is one object, but one object could be two classes at the same time. For instance, a baby girl is both a baby and a girl. 
    \textbf{V}: Vague images where authors had a hard time identifying. 
    }
    \label{fig:c100_all_examples}
\end{figure}

In Figure~\ref{fig:c100_all_examples}, we visualize some of the 4.71\% examples which belong to the CIFAR-100 training set that our AWD trained network doese not fit. Interestingly enough, there are many examples like 0-3, 10-14, 48, and 49 with overlapping classes with one object. 
There are many examples with wrong labels, such as 4-6, 43-45, 28-29, and 51-57.
In many more examples, there are at least two objects in one image, such as 15-25, 28-33, 38-42, and 58-59. 
Unsurprisingly, the model would be better off not fitting such data, as it would only confuse the model to fit the data, which contradicts its already existing and correct conception of the object. 

\subsection{Additional Robustness benefits} \label{subsec:prune_lr} 
\begin{figure}[h]
  % \begin{minipage}[c]{0.50\textwidth}
    \includegraphics[width=\linewidth]{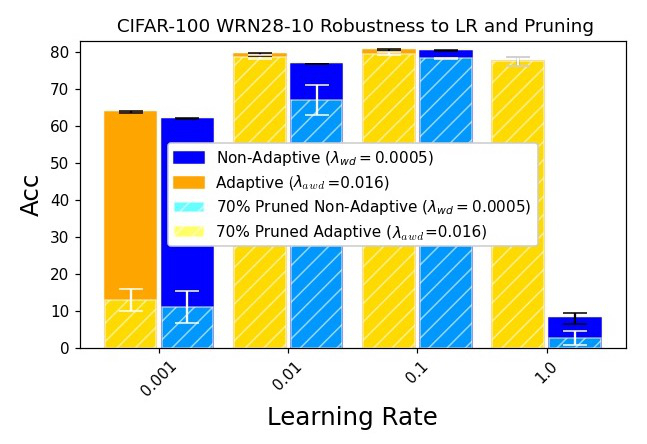}
  % \end{minipage}%\hfill
  % \begin{minipage}[c]{0.50\textwidth}
    % \vspace{-10mm}
    % \hspace{5mm}
    \caption{
       CIFAR-100 models trained with Adaptive Weight Decay (AWD) are less sensitive to learning rate. Also, due to the smaller weight norms of models trained with AWD, they seem like good candidates for pruning. Interestingly, when models are trained with smaller learning rates, they could be more sensitive to trivial pruning algorithms such as non-iterative (i.e., global) $\ell_1$ pruning. The results are average of 4 runs. 
    } \label{fig:condensed_prune_lr_c100}
  % \end{minipage}
  % \vspace{-10mm}
\end{figure}

Throughout the 2D grid search experiments in section~\ref{sec:wd}, we observed that non-adaptive weight decay is sensitive to changes in the learning rate (LR). 
In this section we aim to study the sensitivity of the best hyper-parameter value for adaptive and non-adaptive weight decay to learning rate. 
In addition, models trained with adaptive weight decay tend to have smaller weight norms which could make them more suited for pruning. 
To test this intuition, we adopt a simple non-iterative $\ell_1$ pruning. 
To build confidence on robustness to LR and pruning, for the optimal choices of $\lambda_{wd}=0.0005$ and the estimated ${\lambda_{awd}}=0.016$ for WRN28-10, we train 5 networks per choice of learning rate. 
We prune each network to various degrees of sparsity. 
We then plot the average of all trials per parameter set for each of the methods. 
Figure~\ref{fig:condensed_prune_lr_c100} summarizes the clean accuracy without any pruning and the accuracy after 70\% of the network is pruned. 
As it can be seen, for CIFAR-100, adaptive weight decay is both more robust to learning rate changes and also the result networks are less sensitive to parameter pruning. 
For more details and results on CIFAR-10, please see Appendix~\ref{app:lr} and \ref{app:prune}. 

\subsection{Under-Fitting Data, A Desirable Property} \label{subsec:label}
Our experiments show that adaptive weight decay prevents fitting all the data in case of noisy label and adversarial training. 
Experimentally, we showed that adaptive weight decay contributes to this outcome more than non-adaptive weight decay.
Interestingly, even in the case of natural training of even simple datasets such as CIFAR-100, networks trained with optimal adaptive weight decay still underfit the data. 
For instance, consider the following setup where we train a ResNet50, with ASAM \citep{kwon2021asam} minimizer with both adaptive and non-adaptive weight decay\footnote{To find the best performing hyper-parameters for both settings, we do a 2D grid-search similar to  Figure~\ref{fig:Resnet322DGridsearch}.}. 
Table \ref{tab:asam} shows the accuracy of the two experiments. 

\begin{table*}[h]
	\centering
	\begin{tabular}{rcccc}
    	\toprule
		Method & Learning Rate & $\|W\|_2$ & Test Acc(\%) & Training Acc(\%)  \\ 
    	\midrule
    	Adaptive ${\lambda_{awd}}=0.022$ & 0.01 & 14.39 & 83.21 & 95.29 \\ 
    	Non-Adaptive $\lambda_{wd}=0.005$ & 0.01 &  17.86 & 83.23 & 98.57 \\
		\bottomrule
    \end{tabular}
	\caption{Train and Test accuracy of adaptive and non-adaptive weight decay trained models. 
	While the adaptive version fits 3.28\% less training data, it still results in comparable test accuracy.  
	Our hypothesis is that probably it avoids fitting the noisy labeled data.}
	\label{tab:asam}
\end{table*}

\begin{figure*}[h]
    \centering
    \includegraphics[width=1\textwidth]{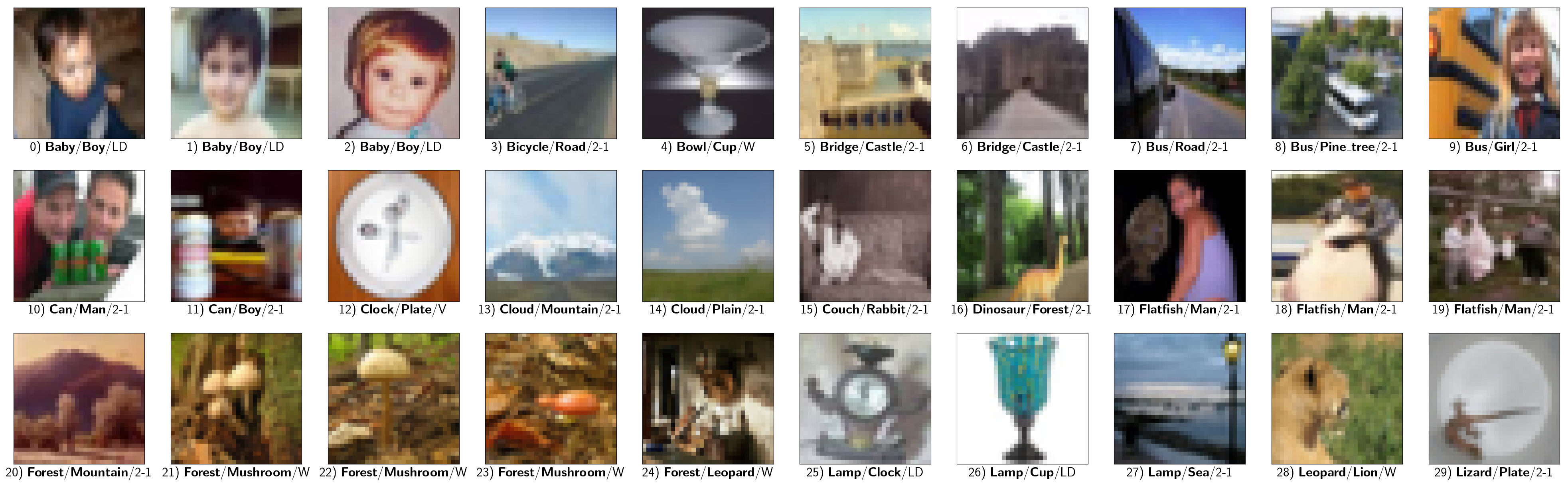}
    \caption{Examples from CIFAR-100 training dataset that have noisy labels. For every image we state [dataset label/ prediction of classifier trained with AWD / our category of noisy case]. We classify these noisy labels into several categories: \textbf{W}: Wrong Labels where the picture is clear enough to comprehend the correct label.; \textbf{2-1}: Two Objects from CIFAR-100 in one image, but only one label is given in the dataset; 
    \textbf{LD}: Loosely Defined Classes where there is one object, but one object could be two classes at the same time. For instance, a baby girl is both a baby and a girl. 
    \textbf{V}: Vague images where authors had a hard time identifying. 
    }
    \label{fig:c100_examples}
\end{figure*}

Intruigingly, without losing any performance, the model trained with adaptive weight decay avoids fitting an extra 3.28\% of training data compared to the model trained with non-adaptive weight decay. 
We investigated more on what the 3.28\% unfit data looks like \footnote{See Appendix~\ref{app:unfit_c100} for images from the 4.71\%.}. 
Previously studies have discovered that some of the examples in the CIFAR-100 dataset have wrong labels \citep{zhang2017method, muller2019identifying, al2018labeling, kuriyama2020autocleansing}. 
We found the 3.28\% to be consistent with their findings and that many of the unfit data have noisy labels. 
We show some apparent noisy labeled examples in Figure~\ref{fig:c100_examples}.

\subsection{Robustness to Sub-Optimal Learning Rate} \label{app:lr}
\begin{figure}[h]
    \centering
    % \subfigure[]{\includegraphics[width=0.49\columnwidth]{new_lr_c10}\label{fig:lr_c10}}
    % \subfigure[]{\includegraphics[width=0.49\columnwidth]{new_lr_c100}\label{fig:lr_c100}}
    \subfigure[]{\includegraphics[width=0.49\columnwidth]{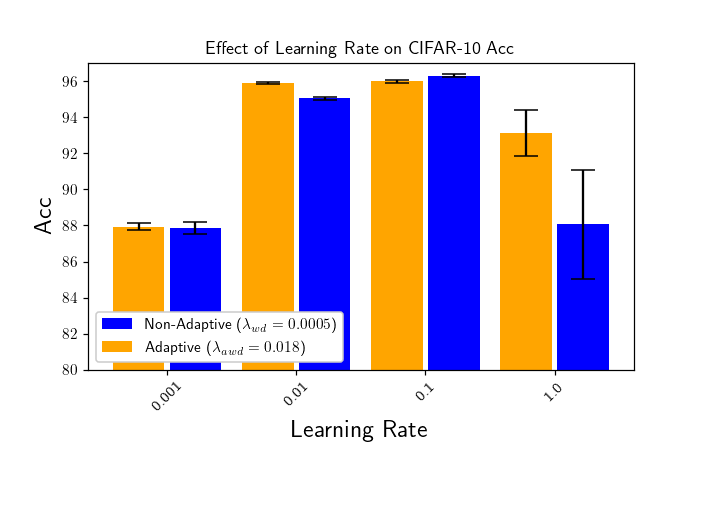}\label{fig:lr_c10}}
    \subfigure[]{\includegraphics[width=0.49\columnwidth]{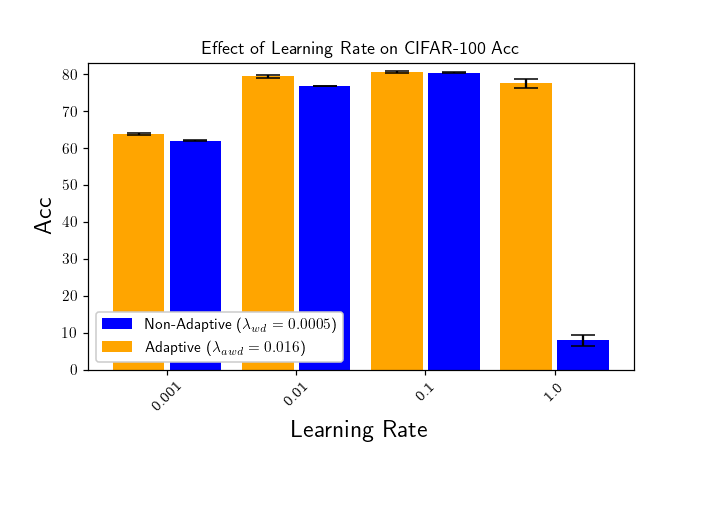}\label{fig:lr_c100}}
    \caption{Models trained with Adaptive Weight Decay are more robust to Learning rate. Results are an average over 5 trials.  
    %We use the same experiment setup and estimated ${\lambda_{awd}}$ values from section~\ref{sec:awd}.
    } 
\end{figure}

In this section, we have a deeper look the performance of networks trained with sub-optimal learning rates. 
We observe that adaptive weight decay is more robust to changes in the learning rate in comparison to non-adaptive weight decay for both CIFAR-10 and CIFAR-100, as shown in Figure \ref{fig:lr_c10} and \ref{fig:lr_c100}. 
The accuracy for $lr=1.0$ for non-adaptive weight decay drops 69.67\% on CIFAR-100 and 5.0\% on CIFAR-10, compared to its adaptive weight decay counterparts.

The robustness to sub-optimal learning rates suggests that adaptive weight decay might be more suitable for applications where tuning for the optimal learning rate might be expensive or impractical. 
An example would be large language models such as GPT-3 or Megatron-Turing NLG, where even training the network once is expensive \citep{brown2020language, smith2022using, rasley2020deepspeed}. 
Another example would be neural architecture search, where one trains many networks \citep{tan2019efficientnet, zhou2018resource, real2019aging, bergstra2013making, mendoza2016towards}.

\subsection{Adaptive weight decay on ImageNet} \label{app:imagenet} 
% \ali{new:}
In this section we illustrate that tuning adaptive weight decay can result in comparable performance to non-adaptive weight decay even in settings where training with non-adaptive weight decay does not suffer from over-fitting. For this purpose we perform free adversarial training \citep{shafahi2019adversarial} with $\epsilon=4/255$ on ImageNet scale. We use a resnet-50 backbone. The parameters used in this setting are replay $m=4$, batchsize of 512, and an initial learning rate of 0.1 which drops by a factor of 0.1 each $n/3$ epochs. 

Similar to the experiments in the main body, we perform a hyper-parameter search to find the best weight decay parameter and the best adaptive weight decay parameter and report both the robustness and clean accuracy of each of the trained models. The results are summarized in Table~\ref{tab:imagenet}.

\begin{table}[h]
    \hspace*{-2em}
	\centering
	\begin{tabular}{rccccc}
    	\toprule
    	Method &                    Model &   $\epsilon$ &  robustness \% &    natural accuracy \%  \\ 
    	\midrule
        $\lambda_{wd} = 0.000004$ &   Resnet-50 & 4 &  26.04 &   55.41 \\
    	$\lambda_{wd} = 0.000005$ &   Resnet-50 & 4 &  26.71 &   54.95 \\ 
        $\lambda_{wd} = 0.000006$ &   Resnet-50 & 4 &  25.84 &   55.13 \\ 
        $\lambda_{wd} = 0.000007$ &   Resnet-50 & 4 &  25.71 &   54.20 \\ 
        \midrule
    	${\lambda_{awd}} = 0.0006$ &   Resnet-50 & 4 &  25.94 &   54.06 \\
        ${\lambda_{awd}} = 0.0007$ &   Resnet-50 & 4 &  \bf{26.84} &   \bf{56.31} \\
        ${\lambda_{awd}} = 0.0008$ &   Resnet-50 & 4 &  26.36 &   54.99 \\
        ${\lambda_{awd}} = 0.0009$ &   Resnet-50 & 4 &  26.39 &   55.69 \\
		\bottomrule
    \end{tabular}
	\caption{Robustness and clean accuracy of resnet-50 models trained with adversarial training for free $m=4$ to be robust against attacks with robustness budget of $\epsilon=4$.}
	\label{tab:imagenet}
\end{table}

\subsection{Robustness to Parameter Pruning} \label{app:prune} 
As seen in previous sections, models trained with adaptive weight decay tend to have smaller weight norms. 
Smaller weight norms, can indicate that models trained with adaptive weight decay are less sensitive to parameter pruning. 
We adopt a simple non-iterative $\ell_1$ pruning to test our intuition. 
Table~\ref{tab:prune} which shows the accuracy of models trained with adaptive and non-adaptive weight decay after various percentages of the parameters have been pruned verifies our hypothesis. 
Adaptive weight decay trained models which are trained with various learning rates are more robust to pruning in comparison to models trained with non-adaptive weight decay.

\begin{figure}[h]
	\centering
	\begin{tabular}{rcccc}
    	\toprule
		\multirow{2}{*}{Method} & \multicolumn{4}{c}{Learning Rate} \\ 
    	& 0.001 & 0.01 & 0.1 & 1.0 \\ 
    	\midrule
    	Adaptive ${\lambda_{awd}}=0.016$ & \bf{92.53} & \bf{16.22} & \bf{18.3} & 26.65 \\
    	Non-Adaptive $\lambda_{wd}=0.0005$ & 116.98 & 27.01 & 24.06 & \bf{12.16} \\
		\bottomrule
    \end{tabular}
	\caption{Norm of the weights for networks trained with adaptive weight decay and non-adaptive weight decay. }
	\label{tab:weight_norm}
\end{figure}

\begin{table}[h]
    \hspace*{-6em}
    \centering
    \setlength{\tabcolsep}{3pt}
	\begin{tabular}{cccccccccc}
    	\toprule
    	Dataset & LR & Method & Nat & 40\% & 50\% & 60\% & 70\% & 80\% & 90\% \\ 
    	\midrule
		\multirow{8}{*}{C10} 
		& \multirow{2}{*}{1}    & ${\lambda_{awd}}=0.018$           & $\bf{93.1\pm1.3}$ & $\bf{93.1\pm1.3}$  & 	$\bf{93.1\pm1.3}$ & 	$\bf{93.1\pm1.3}$ & 	$\bf{93.1\pm1.3}$ & 	$\bf{93.1\pm1.3}$ & 	$\bf{93.2\pm1.3}$ \\ 
		&                           & $\lambda_{wd}=0.0005$ & $88.1\pm3.0$ & $88.0\pm3.1$       & 	$88.0\pm3.1$ & 	$88.0\pm3.1$ & 	$88.0\pm3.1$ & 	$88.0\pm3.1$ & 	$88.0\pm3.1$ \\
        
		& \multirow{2}{*}{0.1}     & ${\lambda_{awd}}=0.018$           & $96.0\pm0.1$ & $91.6\pm8.7$ & 	$91.7\pm8.6$ & 	$91.7\pm8.6$ & 	$91.5\pm9.0$ & 	$90.5\pm10.5$ & 	$83.7\pm17.9$  \\ 
		&                           & $\lambda_{wd}=0.0005$ & $\bf{96.3\pm0.1}$ & $93.6\pm5.3$ & 	$93.6\pm5.3$ & 	$93.6\pm5.3$ & 	$93.4\pm5.5$ & 	$92.6\pm6.2$ & 	$84.3\pm13.6$ \\
		
		& \multirow{2}{*}{0.01}      & ${\lambda_{awd}}=0.018$           & $\bf{95.9\pm0.1}$ & $\bf{95.9\pm0.1}$ & 	$\bf{95.9\pm0.1}$ & 	$\bf{95.9\pm0.1}$ & 	$\bf{95.9\pm0.1}$ & 	$\bf{95.8\pm0.1}$ & 	$\bf{91.4\pm1.4}$ \\ 
		&                           & $\lambda_{wd}=0.0005$ & $95.0\pm0.1$ & $95.0\pm0.1$ & 	$95.0\pm0.1$ & 	$94.5\pm0.3$ & 	$75.2\pm12.9$ & 	$19.3\pm12.6$ & 	$12.2\pm5.0$ \\ 
		
		& \multirow{2}{*}{0.001}        & ${\lambda_{awd}}=0.018$           & $87.9\pm0.2$ & $82.1\pm1.9$ & 	$75.2\pm4.2$ & 	$64.6\pm7.2$ & 	$38.3\pm8.4$ & 	$18.8\pm4.3$ & 	$13.4\pm1.9$ \\ 
		&                           & $\lambda_{wd}=0.0005$ & $87.9\pm0.4$  & $80.5\pm2.9$ & 	$73.8\pm4.8$ & 	$56.4\pm8.1$ & 	$46.8\pm16.9$ & 	$21.0\pm9.9$ & 	$12.2\pm2.5$ \\ 
    	\midrule
    	\multirow{8}{*}{C100} 
		& \multirow{2}{*}{1}    & ${\lambda_{awd}}=0.016$           &   $\bf{77.6\pm1.3}$  & $\bf{77.6\pm1.3}$ & 	$\bf{77.6\pm1.3}$ & 	$\bf{77.6\pm1.3}$ & 	$\bf{77.6\pm1.3}$ & 	$\bf{77.5\pm1.2}$ & 	$\bf{75.3\pm1.3}$ \\ 
		&                           & $\lambda_{wd}=0.0005$ &   $7.9\pm1.4$  & $2.6\pm1.8$ & 	$2.6\pm1.8$ & 	$2.6\pm1.8$ & 	$2.6\pm1.8$ & 	$2.6\pm1.8$ & 	$2.6\pm1.8$ \\
        
		& \multirow{2}{*}{0.1}     & ${\lambda_{awd}}=0.016$           &   $80.7\pm0.3$  & $80.7\pm0.3$ & 	$\bf{80.7\pm0.3}$ & 	$\bf{80.5\pm0.2}$ & 	$\bf{79.6\pm0.4}$ & 	$\bf{74.5\pm0.9}$ & 	$29.2\pm3.6$ \\ 
		&                           & $\lambda_{wd}=0.0005$ &  $80.5\pm0.2$  & $80.4\pm0.1$ & 	$80.2\pm0.2$ & 	$79.9\pm0.2$ & 	$78.4\pm0.2$ & 	$71.3\pm1.2$ & 	$23.9\pm6.7$ \\
		
		& \multirow{2}{*}{0.01}      & ${\lambda_{awd}}=0.016$           & $\bf{79.5\pm0.4}$    & $\bf{79.5\pm0.3}$ & 	$\bf{79.4\pm0.3}$ & 	$\bf{79.3\pm0.4}$ & 	$\bf{78.7\pm0.5}$ & 	$\bf{75.0\pm0.8}$ & 	$\bf{28.9\pm4.4}$ \\ 
		&                           & $\lambda_{wd}=0.0005$ & $76.9\pm 0.1$    & $74.4\pm3.6$ & 	$74.1\pm3.8$ & 	$73.5\pm4.2$ & 	$67.2\pm4.1$ & 	$17.5\pm6.9$ & 	$1.5\pm0.4$  \\ 
		
		& \multirow{2}{*}{0.001}        & ${\lambda_{awd}}=0.016$           & $\bf{63.8\pm0.3}$    & $\bf{55.3\pm1.8}$ & 	$\bf{47.4\pm2.2}$ & 	$\bf{32.0\pm2.1}$ & 	$12.9\pm2.9$ & 	$2.8\pm1.4$ & 	$1.1\pm0.2$ \\ 
	    &                           & $\lambda_{wd}=0.0005$ & $62.1\pm0.2$     & $52.7\pm2.0$ & 	$41.7\pm2.7$ & 	$25.6\pm2.4$ & 	$11.0\pm4.4$ & 	$3.1\pm0.9$ & 	$1.2\pm0.1$ \\ 
		\bottomrule
    \end{tabular}
	\caption{Models trained with Adaptive Weight Decay are more robust to change in learning rate and pruning. }
    \label{tab:prune}
\end{table}

\section{Related Work on Adaptive Weight Decay}\label{app:related_AWD}
As discussed before, AdaDecay \cite{nakamura2019adaptive} and Lars \cite{you2017large} are the methods most related to ours. Here we discuss the major differences between our method and the related works. 

\subsection{AdaDecay} 
The concept of Adaptive Weight Decay was first introduced by \citep{nakamura2019adaptive}.
Similar to our method, their method (AdaDecay) changes the weight decay's hyper-parameter at every iteration. 
Unlike our method, in one iteration, AdaDecay imposes a different penalty to each individual parameter, while our method penalizes all parameters with the same magnitude. 
For instance, let us consider the weight decay updates from eq~\ref{eq:gd_update} for both methods. For parameters $w$ at iteration $t$, the weight decay updates in AWD is $-\lambda_{t} w$, where $\lambda_{t}$ varies at every iteration. 
However, the AdaDecay updates $w_i$ is $ -\lambda \theta_{t, i} w_i$, where $\lambda$ is constant at every iteration, instead, $0 \leq \theta_{t, i} \leq 2$ can vary for different parameters $i$ and for different iterations $t$.
In other words, AdaDecay introduces $\theta_{t, i}$ for every single parameter of the network at every step to represent how strongly each parameter should be penalized. 
For instance, if $ \forall{t,i}: \theta_{t,i}=1$, then AdaDecay has the same effect as traditional non-adaptive weight decay method.

To further understand the differences between AdaDecay and AWD, we build intuition on the values $\theta_{t,i}$ could possibly take. For parameters in layer $L$, Ada decay defines $\theta_{t,i}$ as: 
\begin{equation}
\theta_{t,i} = \frac{2}{1 + exp(-\alpha {\nabla \bar{w}_{i,t}})}    
\end{equation}
where $\bar{\nabla w_{i,t}}$ is the layerwise-normalized gradients.
More precisely, $\nabla \bar{w}_{i,t} = \frac{\nabla w_{i,t} - \mu_{L}}{\sigma_{L}}$ where $\mu_{L}$ and $\sigma_{L}$ represent mean and standard deviation of gradients at layer $L$ respectively. 
Please note that $\mathbb{E}_{w_i \in L} \theta_{t,i} = 1$, due to the fact that $\theta_{t,i}$ is a sigmoid function applied to a distribution with mean of zero and standard deviation of one. In other words, AdaDecay on average does not change the $\lambda_{wd}$ hyper-parameter, while AWD does. In simpler words, AWD can increase the weight norm penalty (i.e. $\lambda_{t}$) indefinitely, while AdaDecay's penalty is bounded. For instance, assuming the most extreme case for AdaDecay where $\nabla \bar{w}_{i,t} = \infty$ for all $i$ and $t$, then $\theta_{t,i}=2$. In other words, the effect of AdaDecay becomes at most twice as strong as non-adaptive weight decay, while our version does not have such upper-bounds.

We also performed experimental evaluations on AdaDecay. The results are summerized in Table~\ref{tab:sota}. We performed a grid-search on the hyper-parameter for AdaDecay and based on PGD-3 adversarial robustness on a held-out validation set, selected the best performing hyper-parameter for AdaDecay. As it can be seen in Table~\ref{tab:sota}, AWD trained network outperforms the AdaDecay trained network by 3\% accuracy. 

\subsection{LARS} \label{app:lars}
LARS \cite{you2017large} is an optimizer that has been widely adopted for large-batch optimization. LARS achieves amazing performance by adaptively changing the learning rate for each layer. 
At first glance, the LARS and AWD seem very similar, where in fact,they are different and can even be applied simultaneously during training.
The similarities raise from the fact that both methods use $\frac{\|w\|_2}{\|\nabla w \|_2}$ as a signal to adaptively change training hyper-parameters, while LARS changes the learning rate and AWD changes the weight decay hyper parameter. 

To better understand the differences between the two methods, let us rewrite eq~\ref{eq:gd_update} for LARS. At step $t+1$ for a network with only one layer with parameter $w$, we will have:

\begin{equation}
  w_{t+1} = w_t - lr \times \nu \times \frac{ \| w \|}{\|\nabla w\| + \lambda \|w\|} ( w + \nabla w)  
\end{equation}

Note that in this setting, the ratio between the updates from the main loss (i.e., $\nabla w$) and weight decay(i.e., $w$) is untouched, while AWD enforces stronger regularization effects by breaking this constant ratio. Applying AWD we have:

\begin{equation}
  w_{t+1} = w_t - lr \times (\nabla w + \lambda \frac{\| \nabla w \|}{\| w \|} w)  
\end{equation}

Also, note that since the two methods are independent of one another, they can be combined. So using AWD and LARS together, we will have:

\begin{equation}
  w_{t+1} = w_t - lr \times \nu \times \frac{ \| w \|}{\|\nabla w\|} (\lambda \frac{\| \nabla w \|}{\| w \|} w + \nabla w)  
\end{equation}

To experimentally study the differences between the two methods, we repeat the experiments done for Table \ref{tab:nat_perf_adv_trained} with and without the LARS optimizer on a ResNet18 architecture. For LARS hyper-parameters, we use trust-coefficient=$0.02$, $\epsilon=1e-8$, and we clip the gradients. Table \ref{tab:lars} summarizes the results of these experiments.

\begin{table*}[h]
    % \vspace{-1.em}
	\centering
	\begin{tabular}{rccccccc}
    	\toprule
    	Method &						Dataset&	Opt & 	$\|W\|_2$ & 	Nat Acc &		AutoAtt & $Xent + \frac{\lambda_{wd}^* \cdot \|W\|_2^2}{2}$ \\     	
        \midrule
        
    	$\lambda_{wd} = 0.00089$ 	&	\multirow{4}{*}{CIFAR-10} &	SGD & 			28.42 & 		83.43 &  		43.20 & 1.11 \\ 
    	${\lambda_{awd}}=0.01834$ &		&  SGD & 			\textbf{7.89} & 		\textbf{85.99} &			\textbf{46.87} & \textbf{0.80} \\ 
        $\lambda_{wd} = 0.00089$ 	& &	LARS & 			27.87 & 		83.79 &  		43.28 & 1.11 \\ 
    	${\lambda_{awd}}=0.02181$ &		& LARS & 	\textbf{6.20} & 		\textbf{84.74} &			\textbf{46.82} & \textbf{0.99} \\
    	\midrule
        $\lambda_{wd} = 0.0.00211$ 	&	\multirow{4}{*}{CIFAR-100} &	SGD & 			22.69 & 		58.04 &  		21.94 & 2.43 \\ 
    	${\lambda_{awd}}=0.02181$ &		&  SGD & 			\textbf{12.89} & 		\textbf{58.46} &			\textbf{24.98} & \textbf{2.32} \\ 
        $\lambda_{wd} = 0.00211$ 	& &	LARS & 			22.91 & 		57.82 &  		21.72 & 2.40 \\ 
    	${\lambda_{awd}}=0.02181$ &		& LARS & 	\textbf{13.11} & 		\textbf{58.90} &			\textbf{24.89} & \textbf{2.23} \\
		\bottomrule
    \end{tabular}
	\caption{Adversarial robustness of PGD-7 adversarially trained ResNet18 on CIFAR10 and CIFAR100 datasets. Similar to the Table \ref{tab:nat_perf_adv_trained}, we used a grid search to find the best $\lambda_{wd}$ hyper-parameter for non-adaptive rows and ${\lambda_{awd}}$ for adaptive rows. The table compares the effect of AWD and LARS. As can be seen, the AWD trained models outperform the non-adaptive counter-parts in terms of robustness, natural accuracy, smaller weight norms, and smaller non-adaptive training loss. Interestingly enough, LARS does not have any major effects on any of the mentioned metrics, unless combined with AWD. }%\ali{caption needs to be updated -- also ??? should be replaced with correct values}}
	\label{tab:lars}
\end{table*}

\section{Limitations}
% TODO:  We need to discuss the following: 
As discussed throughout the paper, AWD is most effective when overfitting happens. In other words, AWD would not be effective in settings that we underfit the training data such as ImageNet dataset. Needless to say, in such settings, AWD does not hurt the accuracy either. Section~\ref{app:imagenet} summerizes our ImageNet results for the adversarial robustness setting. As it can be seen in Table~\ref{tab:imagenet}, we gain minor improvements in terms of adversarial robustness by adopting AWD.

Similar to the traditional weight decay, AWD requires hyper-parameter tuning. Our experiments show that AWD works best if its hyper-parameter is tuned per dataset and per network architecture.

\section{An Example for Convergence of Adaptive Weight Decay}
As discussed previously in Section \ref{sec:awd_diff_trad}, we treat the weight decay hyper-parameter computed in each iteration as a non-tensor constant and we do not let the gradients to back-propagate through the computation of $\lambda_{wd(t)}$ in eq.~\ref{eq:awd}. 
So in some sense, every optimization step of the adaptive weight decay is just an optimization step in the traditional non-adaptive weight decay, with the only difference that the weight decay hyper-parameter is being scaled.
Consequently, in terms of convergence, it is not unlikely that adaptive and non-adaptive weight decay have fairly similar behavior. 

In this section, we provide a mathematical demonstration for the convergence of AWD for a very simple convex optimization problem. 
Consider the following optimization problem: 
\begin{equation}
 \min_{x} MSE(x, \beta) = \min_{x} \frac{\|x-\beta\|^2}{2} \label{eq:simple_convex}
\end{equation}
This problem is convex and even has a closed-form solution. 
However, let us consider the SGD solution with an adaptive weight decay regularizer: 
\begin{equation}
 \min_{x} MSE(x, \beta) = \min_{x} \frac{\|x-\beta\|^2}{2} + c \frac{|x - \beta|}{|x|} \label{eq:simple_convex}
\end{equation}
where, $c = \frac{\lambda_{awd}}{2}$ and $x \neq 0$. 
We consider all cases for possible choices of $\beta$ and $x$ and show that for all 7 cases, as long as we chose the hyper-parameter $0 < c < \frac{\beta^2}{2}$, which translates to $0 < \lambda_{awd} < \beta^2$, the problem is locally convex in those regimes. 
In order to do that, we show that the second derivative is always positive in all cases. 
Table \ref{tab:theory_proof} summarizes all 7 cases and the conditions in which the second derivative is guaranteed to be positive.

\begin{table*}[t]
	\centering
    \hspace*{-4em}
	\begin{tabular}{rccccccc}
    	\toprule
    	$\beta$ & $x$ & $\beta$ vs. $x$ & Minimization & Simplified Minimization & 2nd Deriv. & Conv. Cond. & $c$ Cond.\\
        \midrule
        $\beta>0$ & $x>0$ & $x>\beta$ & $0.5\|x-\beta\|^2+c\frac{ \|x-\beta\| }{ \|x\| }$ & $0.5(x-\beta)^2+c\frac{x-\beta}{x}$ & $1-\frac{2bc}{x^3}$ & $1-\frac{2bc}{x^3}>0$ & $c<\frac{b^2}{2}$\\
$\beta>0$ & $x>0$ & $x<\beta$ & $0.5\|x-\beta\|^2+c\frac{ \|x-\beta\| }{ \|x\| }$ & $0.5(x-\beta)^2-c\frac{x-\beta}{x}$ & $1+\frac{2bc}{x^3}$ & $1+\frac{2bc}{x^3}>0$ & $c>0$\\
$\beta>0$ & $x<0$ & $x<\beta$ & $0.5\|x-\beta\|^2+c\frac{ \|x-\beta\| }{ \|x\| }$ & $0.5(x-\beta)^2+c\frac{x-\beta}{x}$ & $1-\frac{2bc}{x^3}$ & $1-\frac{2bc}{x^3}>0$ & $c>0$\\
$\beta<0$ & $x>0$ & $x>\beta$ & $0.5\|x-\beta\|^2+c\frac{ \|x-\beta\| }{ \|x\| }$ & $0.5(x-\beta)^2+c\frac{x-\beta}{x}$ & $1-\frac{2bc}{x^3}$ & $1-\frac{2bc}{x^3}>0$ & $c>0$\\
$\beta<0$ & $x<0$ & $x>\beta$ & $0.5\|x-\beta\|^2+c\frac{ \|x-\beta\| }{ \|x\| }$ & $0.5(x-\beta)^2-c\frac{x-\beta}{x}$ & $1+\frac{2bc}{x^3}$ & $1+\frac{2bc}{x^3}>0$ & $c>0$\\
$\beta<0$ & $x<0$ & $x<\beta$ & $0.5\|x-\beta\|^2+c\frac{ \|x-\beta\| }{ \|x\| }$ & $0.5(x-\beta)^2+c\frac{x-\beta}{x}$ & $1-\frac{2bc}{x^3}$ & $1-\frac{2bc}{x^3}>0$ & $c<\frac{b^2}{2}$\\
$\beta=0$ & Any & - & $0.5\|x-\beta\|^2+c\frac{ \|x-\beta\| }{ \|x\| }$ & $(0.5x-\beta)^2$ & 1 & Always convex & Any $c$\\
		\bottomrule
    \end{tabular}
	\caption{ Conditions guaranteeing convexity and convergence of AWD for a simple regression problem.}
	\label{tab:theory_proof}
\end{table*}

As it can be seen from Table \ref{tab:theory_proof}, $0 < c < \frac{\beta^2}{2}$ satisfies the local convexity condition for all 7 cases. 
As a result, adaptive weight decay formulation for this problem is always locally convex and given the right hyper-parameter for $\lambda_{awd}$, SGD always converges to a solution. 

\end{document}